\documentclass{article}
\usepackage{iclr2027_conference,times}

\usepackage{amsmath,amsfonts,bm}

\def\eqref#1{equation~\ref{#1}}

\def\1{\bm{1}}

\DeclareMathAlphabet{\mathsfit}{\encodingdefault}{\sfdefault}{m}{sl}
\SetMathAlphabet{\mathsfit}{bold}{\encodingdefault}{\sfdefault}{bx}{n}

\usepackage{latexsym}
\usepackage{amsmath}
\usepackage{amsfonts}
\usepackage{amssymb}
\usepackage{booktabs}
\usepackage{array}
\usepackage{multirow}
\usepackage{makecell}
\usepackage{adjustbox}
\usepackage{caption}
\usepackage{graphicx}
\usepackage{placeins}
\usepackage{float}
\usepackage{listings}
\usepackage[table]{xcolor}
\usepackage{enumitem}
\usepackage{xurl}
\usepackage{hyperref}
\usepackage{url}
\usepackage[most]{tcolorbox}
\usepackage{xcolor}
\IfFileExists{fontawesome5.sty}{
    \usepackage{fontawesome5}
}{
    \newcommand{\faCog}{\texttt{[S]}}
    
}

\definecolor{systemColor}{RGB}{230, 240, 255}
\definecolor{inputColor}{RGB}{255, 253, 235}
\definecolor{borderColor}{RGB}{200, 200, 200}
\definecolor{annotationColor}{RGB}{247,249,244}

\definecolor{TableNavy}{RGB}{35, 73, 108}
\definecolor{TableBlue}{RGB}{73, 122, 164}
\definecolor{TableRule}{RGB}{132, 157, 179}
\definecolor{TableSoft}{RGB}{95, 122, 145}
\definecolor{TableHighlight}{RGB}{238, 246, 252}
\definecolor{TableSummary}{RGB}{245, 249, 252}
\newcommand{\best}[1]{\textbf{{#1}}}
\newcommand{\second}[1]{\textcolor{TableBlue}{\underline{#1}}}
\newcommand{\keylabel}[1]{\textbf{{#1}}}
\newcommand{\sectiontag}[1]{\textbf{\textcolor{TableSoft}{\textsc{#1}}}}
\newcommand{\ourscell}[1]{\textbf{{#1}}}
\newcommand{\metricup}[1]{\textbf{#1}\hspace{1pt}\textcolor{TableBlue}{\scriptsize$\uparrow$}}
\newcommand{\metricdown}[1]{\textbf{#1}\hspace{1pt}\textcolor{TableBlue}{\scriptsize$\downarrow$}}
\newcommand{\theadrow}{}
\newcommand{\subheadrow}{}
\newcommand{\highlightrow}{\rowcolor{TableHighlight}}
\newcommand{\summaryrow}{\rowcolor{TableSummary}}
\arrayrulecolor{TableRule}
\newcommand{\TableTopRule}{\arrayrulecolor{TableNavy}\specialrule{0.90pt}{0pt}{1.0pt}\arrayrulecolor{TableRule}}
\newcommand{\TableBottomRule}{\arrayrulecolor{TableNavy}\specialrule{0.90pt}{1.0pt}{0pt}\arrayrulecolor{TableRule}}
\newcommand{\TableGroupRule}{\arrayrulecolor{TableRule}\specialrule{0.42pt}{0.7pt}{0.7pt}}

\tcbset{
    promptbox/.style={
        enhanced,
        boxrule=0.5pt,
        colframe=borderColor,
        left=4pt,
        right=4pt,
        top=8pt,
        bottom=2pt,
        before skip=4pt,
        after skip=4pt,
        lines before break=2,
        fonttitle=\bfseries\small\sffamily,
        coltitle=black,
        attach boxed title to top left={
            yshift=-10pt,
            xshift=10pt
        },
        boxed title style={
            colback=white,
            frame hidden
        },
        breakable,
        arc=2pt
    }
}

\newtcolorbox{systembox}[1][]{
    promptbox,
    colback=TableHighlight,
    title=\faCog\ System Prompt,
    #1
}

\newtcolorbox{annotationbox}[1]{
    promptbox,
    colback=annotationColor,
    title=#1
}

\newtcolorbox{taskbox}[1]{
    promptbox,
    colback=inputColor,
    title=Task ID \nolinkurl{#1}
}

\graphicspath{{figs/}{figs/}{./}}
\hypersetup{
    colorlinks=true,
    citecolor=blue,
    linkcolor=blue,
    urlcolor=blue
}

\title{LocalLSTC: A Long Short-Term Control Architecture for Locally Deployed GUI Agents}

\author{
Weiming Li, Helen Paik \& Yulei Sui\thanks{Corresponding author: \texttt{y.sui@unsw.edu.au}} \\
School of Computer Science and Engineering \\
The University of New South Wales \\
Sydney, Australia \\
\texttt{\{weiming.li1,h.paik,y.sui\}@unsw.edu.au}
}

\iclrfinalcopy
\begin{document}

\maketitle

\begin{abstract}
Modern GUI-agent frameworks achieve strong desktop task performance with frontier API models, yet persistent control information often remains implicit in growing interaction trajectories. At each step, the planner reconstructs the active task stage, accumulated evidence, and runtime feedback before deciding the next action. This dependence becomes more pronounced under weaker local reasoning backbones. Across four representative state-of-the-art frameworks, replacing GPT-5 with Qwen3.5-9B reduces average OSWorld SR-100 from 60.9\% to 35.2\%. Trajectory annotation further identifies at least one control failure in 94.7\% of failed trajectories. To address this problem, we introduce LocalLSTC, a training-free architecture that organizes control by temporal scope, maintaining persistent cross-step state to guide short-term execution commitments. The framework comprises two complementary mechanisms. Long-to-Short Planning forms each commitment from persistent state, while Short-to-Long Control integrates execution outcomes back into that state for progress assessment, recovery, and termination. With Qwen3.8-27B, LocalLSTC reaches 73.7\% SR-100 on OSWorld and 68.5\% on WindowsAgentArena, achieving the strongest local OSWorld result and a new state of the art on WindowsAgentArena. Ablations further support contributions from both mechanisms. Together, these results show that temporal organization of control information can reduce cross-step control failures and complement backbone scaling in locally deployed GUI agents.
\end{abstract}

\section{Introduction}

\begin{figure*}[t!]
\centering
\includegraphics[width=\textwidth]{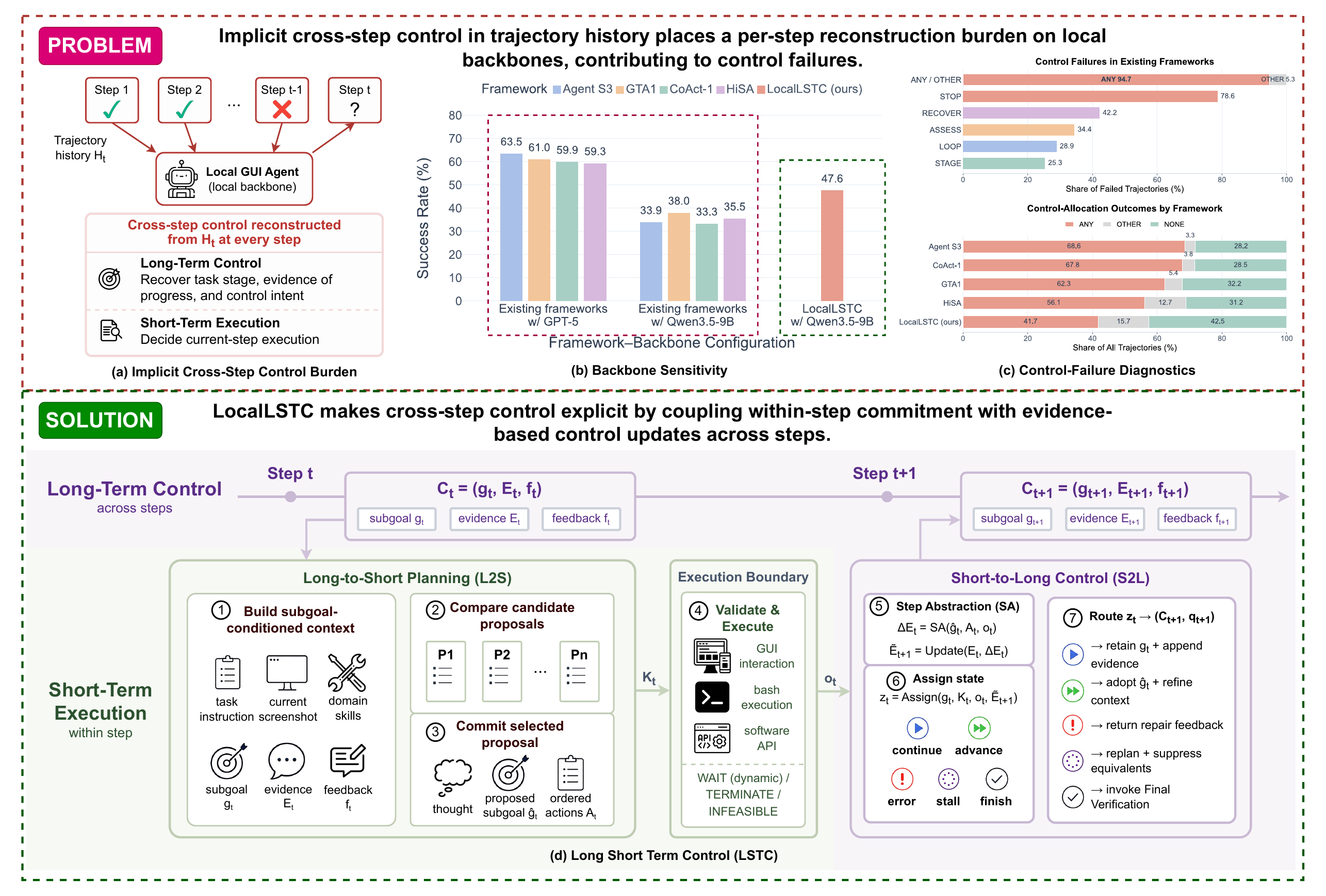}
\caption{Overview of the control problem, empirical evidence, and LocalLSTC. Panel~(a) illustrates the implicit trajectory-level control burden. Panel~(b) shows the SR-100 drop after replacing GPT-5 with Qwen3.5-9B across four GUI-agent frameworks. Panel~(c) reports cross-step control-failure incidence under Qwen3.5-9B. Panel~(d) presents the seven-step L2S--S2L control loop.}
\label{fig:core-idea}
\end{figure*}

GUI agents extend multimodal models from interface understanding to sequential action in computer-use tasks. Long-horizon tasks require agents to preserve the active task stage across interactions, interpret execution outcomes, recover from failures, and terminate only after task constraints are satisfied. Recent systems improve desktop task performance through hierarchical planning, specialized components, code execution, and test-time search \citep{agashe2024agents,agashe2025agents2,gonzalez2025scaling,yang2025gta1,song2025coact}. Yet persistent control information often remains implicit in the growing interaction trajectory. At each decision, the planner needs to reconstruct the active task stage, accumulated evidence, and relevant runtime feedback before determining how to proceed. We call this repeated reconstruction the \textit{implicit trajectory-level control burden}. As illustrated in Figure~\ref{fig:core-idea}(a), reconstruction errors can manifest as loss of stage commitment, incorrect progress assessment, repeated no-progress behavior, mismatched recovery, or premature termination. We refer to these observable manifestations as \textit{control failures}.

Local deployment makes this control burden particularly consequential. Keeping inference local allows screenshots, files, credentials, browser state, and application data to remain within the local deployment boundary, while reducing dependence on external-service latency, availability, and recurring invocation costs \citep{liu2024mobilellm,xue2024powerinfer2}. However, public benchmarks and open-model studies continue to show a substantial capability gap between locally deployed open models and strong API models on desktop tasks \citep{xie2024osworld,bonatti2024windowsagentarena,wang2025opencua}. Recent reasoning models narrow this gap \citep{qwen3.5,qwen3.6-27b,qwen3.8,googledeepmind2026gemma4}, but deployable model scale remains constrained by local hardware capacity, compute, and inference throughput \citep{liu2024mobilellm,xue2024powerinfer2}. Existing approaches primarily strengthen model parameters or system composition through GUI-specific training, specialized components, or additional inference-time computation \citep{agashe2025agents2,song2025coact,qin2025uitars,xu2024aguvis}. Runtime interfaces for adapting off-the-shelf general backbones to long-horizon desktop control remain less systematically studied \citep{xu2026lifeharness}.

Across four representative state-of-the-art GUI-agent frameworks, our Qwen3.5-9B evaluations average 35.2\% OSWorld SR-100, 25.7 percentage points below the reported GPT-5 average of 60.9\%, as shown in Figure~\ref{fig:core-idea}(b). Averaged equally across the four frameworks, 94.7\% of failed local-backbone trajectories contain at least one control failure. Figure~\ref{fig:core-idea}(c) shows that these failures span stage maintenance, progress assessment, repeated no-progress behavior, recovery, and termination, with the dominant category varying across frameworks.

To address this problem, we introduce LocalLSTC, a Long Short-Term Control Architecture for Locally Deployed GUI Agents. LocalLSTC organizes control by temporal scope, maintaining persistent cross-step state to guide bounded short-term execution commitments. The framework comprises two complementary mechanisms. Long-to-Short Planning (L2S) forms each commitment from the persistent state and current observation, while Short-to-Long Control (S2L) integrates execution outcomes back into that state for subsequent progress assessment, recovery, and termination. The resulting L2S--S2L loop, illustrated in Figure~\ref{fig:core-idea}(d), carries the active subgoal, accumulated evidence, and runtime feedback across interactions without requiring each planner call to reconstruct them from the full trajectory.

Empirically, LocalLSTC improves both task performance and cross-step control behavior under constrained local backbones. As shown in Figure~\ref{fig:core-idea}(b), under the same Qwen3.5-9B backbone, LocalLSTC reaches 47.6\% SR-100, outperforming the four-framework baseline average of 35.2\% by 12.4 percentage points. Figure~\ref{fig:core-idea}(c) further shows that LocalLSTC reduces \textsc{Any} control-failure incidence from the 63.7\% baseline average to 41.7\%. With Qwen3.8-27B, LocalLSTC reaches 73.7\% SR-100 on OSWorld and 68.5\% on WindowsAgentArena, achieving the strongest local OSWorld result and a new state of the art on WindowsAgentArena. Fixed-backbone ablations further show contributions from both L2S and S2L. Together, these results indicate that explicitly organizing control information across temporal scopes can reduce cross-step control failures and complement backbone scaling in locally deployed GUI agents. Our contributions are as follows.
\begin{itemize}[leftmargin=*, nosep]
\item We quantify reasoning-backbone sensitivity across four state-of-the-art GUI-agent frameworks and characterize failed local trajectories with five observable cross-step control-failure labels.
\item We introduce LocalLSTC, a training-free architecture that organizes persistent cross-step state and bounded short-term execution through complementary L2S and S2L mechanisms.
\item LocalLSTC achieves the strongest local results on OSWorld and establishes a new state of the art on WindowsAgentArena. Under a fixed backbone, LocalLSTC shows fewer cross-step control failures than the baselines, while ablations support contributions beyond backbone scaling alone.
\end{itemize}

\section{Related Work}

\subsection{GUI-Specific Models and Parameter Adaptation}

One route to improving locally deployable GUI agents is to strengthen the model itself through GUI-specific data and parameter adaptation. UI-TARS and Aguvis jointly train grounding, reasoning, and GUI action generation \citep{qin2025uitars,xu2024aguvis}. OS-ATLAS scales GUI grounding data to improve localization and out-of-distribution generalization, while AgentCPM-GUI combines grounding-aware pre-training, trajectory supervised fine-tuning, and reinforcement fine-tuning for mobile interaction \citep{wu2024osatlas,zhang2025agentcpmgui}. These approaches improve GUI competence by encoding task-relevant perception, grounding, and interaction patterns into model parameters, but their gains depend on additional training data and adaptation procedures. LocalLSTC operates with an off-the-shelf backbone and leaves model parameters unchanged, providing a training-free control alternative for improving long-horizon GUI interaction.

\subsection{GUI Agent Frameworks and Inference-Time Scaling}

Beyond parameter adaptation, GUI-agent frameworks improve performance by strengthening the system around the backbone through composition, execution tools, and additional inference-time computation. Agent S and Agent S2 combine hierarchical planning, grounding, and execution through an Agent-Computer Interface and generalist--specialist coordination \citep{agashe2024agents,agashe2025agents2}. Agent S3 and GTA1 increase test-time computation through candidate generation, rollout selection, and judge-based ranking \citep{gonzalez2025scaling,yang2025gta1}. CoAct-1 extends GUI interaction with programmatic execution, while Avenir-Web and ClawGUI develop grounding and system infrastructure for broader computer-use settings \citep{song2025coact,li2026avenirweb,tang2026clawgui}. These systems demonstrate that stronger orchestration, richer execution interfaces, and additional inference-time computation can substantially improve computer-use performance. LocalLSTC is compatible with these system-level enhancements and contributes a complementary control structure for maintaining task state and execution evidence across interactions.

\subsection{Sequential Planning and Runtime Control}

Beyond these system-level approaches, sequential agents organize long-horizon decisions through search, subgoals, explicit execution states, and revisable plans. Language Agent Tree Search explores candidate trajectories with inference-time search, while the Subgoal-driven Framework tracks task progress through online subgoal planning \citep{zhou2023lats,wang2026subgoal}. Agent-SAMA uses a finite-state machine for execution verification and recovery, AgentProg represents interaction history as programs with variables and control flow, and Plover exposes plans as inspectable and revisable objects \citep{guo2026agentsama,tian2025agentprog,venkatesan2026plover}. LIFE-HARNESS adapts a runtime harness around a fixed model \citep{xu2026lifeharness}. HiSA introduces hierarchical state abstraction for GUI agents through Step Abstraction and periodic Context Refinement \citep{li2026hisa}. LocalLSTC adopts these evidence-compression mechanisms while coupling them to an L2S--S2L loop that organizes persistent state and bounded execution by temporal scope.

\section{Method}
\label{sec:method}

\subsection{Problem Formulation and State Representation}

A common GUI-agent control interface conditions each decision on the current observation and accumulated interaction trajectory. At step $t$,

$$
a_t=\pi(x_t,H_t), \qquad
o_t=\operatorname{Exec}(a_t), \qquad
H_{t+1}=H_t\oplus(a_t,o_t),
$$

where $x_t$ contains the task instruction and current observation, $H_t$ is the interaction trajectory accumulated through step $t$, and $a_t$ is the action selected by the policy $\pi$. The execution operator $\operatorname{Exec}$ applies $a_t$ to the environment and returns the resulting outcome $o_t$. The operator $\oplus$ appends the current action--outcome pair $(a_t,o_t)$ to the trajectory, producing $H_{t+1}$. This trajectory-conditioned interface leaves the active task stage, accumulated evidence, and runtime feedback implicit in $H_t$, requiring them to be reconstructed before each new decision.

LocalLSTC externalizes information that persists across interactions as a \textit{Long-Term Control State}
$$
C_t=(g_t,E_t,f_t), \qquad
C_0=(\bot,\varnothing,\bot).
$$

where $g_t$ represents the active subgoal and unmet constraints, $E_t$ records subgoal-aligned execution evidence, and $f_t$ carries runtime feedback from errors, stalls, or rejected termination. Together, these variables preserve task stage, progress evidence, and control feedback across planner calls.

The corresponding step-local object is a \textit{Short-Term Execution Commitment}
\[
K_t=(\hat{g}_t,A_t),
\]
where $\hat{g}_t$ is the execution subgoal selected for the current step and $A_t$ is the ordered action list generated under that subgoal. 

L2S forms each Short-Term Execution Commitment from persistent state and the current observation, while S2L incorporates its outcome into the next Long-Term Control State. The interaction between the two temporal scopes is

$$
K_t=\operatorname{L2S}(x_t,C_t), \qquad
o_t=\operatorname{Exec}(A_t), \qquad
(C_{t+1},q_{t+1})=\operatorname{S2L}(C_t,K_t,o_t),
$$

where L2S maps observation $x_t$ and persistent state $C_t$ to commitment $K_t$. Executing its action list $A_t$ produces $o_t$. If execution halts early, $o_t$ records only the executed prefix and its outcomes. The unexecuted suffix provides no evidence. S2L incorporates the commitment and outcome into $C_{t+1}$ and determines task status $q_{t+1}$. Here, $q_t\in\{\texttt{running},\texttt{success},\texttt{failure}\}$, with $q_0=\texttt{running}$. LocalLSTC is instantiated with a planner backbone and an optional visual grounder, which supplies GUI coordinates when the planner has insufficient visual grounding capability.

\subsection{L2S--S2L Control Loop}

Panel~(d) of Figure~\ref{fig:core-idea} instantiates the interaction between the two temporal scopes as a seven-step control cycle. Steps~1--3 form the Short-Term Execution Commitment $K_t$ from the current observation and persistent state. Step~4 validates and executes the commitment to produce outcome $o_t$. Steps~5--7 abstract the execution outcome, determine the corresponding control transition, and update the Long-Term Control State $C_{t+1}$ and task status $q_{t+1}$. The updated state then conditions the next interaction. The following subsections describe each part of the L2S--S2L control loop in detail.

\subsubsection{Long-to-Short Planning}
\label{sec-long-to-short-planning}

L2S converts persistent control state into the execution commitment for the current step. In step~1, $g_t$ provides the active stage reference, $E_t$ provides evidence accumulated from prior execution, and $f_t$ provides feedback from the preceding transition. The initial call derives the first subgoal from the task instruction and current observation. Subsequent calls condition planning on the persisted state. Domain skills are selected from the task-related applications and focused by the foreground application when multiple applications are involved, keeping operational guidance aligned with the active execution context. Appendix~\ref{app:planner-context-interface} details planner context construction.

In step~2, the planner may generate one to three candidate proposals as sequential autoregressive text within a single completion, with all candidates conditioned on the same persistent control state. The same completion then emits a \texttt{Best} candidate index and a brief selection rationale before producing the final executable JSON. No additional candidate sampling, planner call, or separate judge is used. This differs from inference-time approaches that sample candidates across multiple planner calls and rank them with a separate judge, as in GTA1 \citep{yang2025gta1}. Appendix~\ref{app:planner-format-prompt} provides the response format.

In step~3, the selected proposal is serialized as a JSON object containing \texttt{thought}, \texttt{subgoal}, and \texttt{actions}. The proposal specifies the execution subgoal $\hat{g}_t$, where the reserved value \texttt{continue} retains $g_t$ and any other value proposes a new subgoal, together with an action list $A_t$. Actions can combine GUI, Bash, and software API channels as needed within the same commitment. LocalLSTC adopts programmatic Bash execution from CoAct-1 \citep{song2025coact}, while its software API reuses the schema-constrained interface and execution handlers from ComputerRL \citep{lai2025computerrl}. Appendix~\ref{app:planner-context-interface} specifies the planner interface and action space.

\subsubsection{Execution}

Step~4 validates and executes $K_t=(\hat{g}_t,A_t)$. Valid actions are dispatched in list order through the corresponding executor. Reserved actions \texttt{WAIT}, \texttt{TERMINATE}, and \texttt{INFEASIBLE} are handled directly by the runtime. The resulting outcome $o_t$, together with $K_t$, are passed to S2L. Execution is supported by deterministic runtime operations for output parsing, schema validation, action fingerprinting, and timeout handling, with the corresponding runtime details provided in Appendix~\ref{app:implementation-details}.

\subsubsection{Short-to-Long Control}
\label{sec-short-to-long-control}

S2L converts the outcome of the current commitment into the next persistent control state. In step~5, Step Abstraction (SA) evaluates the observed execution outcome against the intended effect of the current commitment. Given the execution subgoal $\hat{g}_t$, ordered actions $A_t$, and observed outcome $o_t$, Step Abstraction determines whether execution produces task-relevant progress, no change, or an exception, and records this assessment as a subgoal-aligned evidence increment
\[
\Delta E_t=\operatorname{SA}(\hat{g}_t,A_t,o_t), \qquad
\widetilde{E}_{t+1}=\operatorname{Update}(E_t,\Delta E_t).
\]
where $\Delta E_t$ denotes the subgoal-aligned evidence extracted from the current execution outcome, and $\widetilde{E}_{t+1}$ is the evidence state after incorporating this increment into $E_t$. This process converts heterogeneous execution traces into evidence interpreted with respect to the action intent and current subgoal. The mechanism builds on HiSA's Step Abstraction for GUI execution and extends the evidence interface to Bash and software API channels \citep{li2026hisa}. Appendix~\ref{app:step-abstraction-details} details the channel-specific observation and abstraction mechanisms.

In step~6, the runtime assigns a transient routing state

$$
z_t=\operatorname{Assign}(g_t,K_t,o_t,\widetilde{E}_{t+1}),
$$

where $z_t\in\{\texttt{continue},\texttt{advance},\texttt{error},\texttt{stall},\texttt{finish}\}$. Assignment follows a fixed precedence. A \texttt{TERMINATE} request maps to \texttt{finish}. Other execution failures map to \texttt{error}. Repeated no-progress maps to \texttt{stall} only when it remains under the current subgoal. A proposed execution subgoal different from the persistent subgoal maps to \texttt{advance}. All remaining cases map to \texttt{continue}. Appendix~\ref{app:state-control-details} specifies the detection rules and thresholds.

In step~7, the routing state determines the next control state
\[
(C_{t+1},q_{t+1})
=\operatorname{Route}(z_t,C_t,K_t,\widetilde{E}_{t+1}).
\]
where $\operatorname{Route}$ applies the transition specified by routing state $z_t$ to the previous control state $C_t$, the current commitment $K_t$, and the updated evidence $\widetilde{E}_{t+1}$, producing the next control state $C_{t+1}$ and task status $q_{t+1}$. The \texttt{continue} route retains the current subgoal state and appends new evidence, while the \texttt{advance} route adopts the proposed subgoal state. Successful \texttt{continue} and \texttt{advance} transitions clear stale runtime feedback. An \texttt{advance} can trigger Context Refinement, which compresses accumulated evidence into compact history while retaining recent evidence for subsequent planning and verification (Appendix~\ref{app:context-refinement-details}). LocalLSTC inherits this refinement mechanism from HiSA but replaces its fixed-period schedule with execution-state triggering \citep{li2026hisa}. The \texttt{error} route writes execution-specific feedback into the persistent state, while \texttt{stall} records no-progress feedback for the next planner call. The \texttt{finish} route invokes Final Verification, which evaluates task completion against accumulated visual, textual, and software API evidence (Appendix~\ref{app:verification-details}). A \texttt{PASS} sets the task status to \texttt{success}. A \texttt{FAIL} records the rejected termination as a \texttt{stall}, writes recovery feedback into the persistent state, and resumes planning with status \texttt{running}. The \texttt{INFEASIBLE} action and predefined runtime failure boundaries set $q_{t+1}=\texttt{failure}$. Step Abstraction, Context Refinement, and Final Verification use the planner backbone under dedicated prompts (Appendix~\ref{app:prompts}), while state assignment and routing are deterministic runtime operations.

\section{Experiments}
\label{sec:experiments}

Experiments evaluate three aspects of LocalLSTC. We first examine planner-backbone sensitivity and cross-step control failures in existing GUI-agent frameworks. We then evaluate LocalLSTC across two desktop benchmarks and multiple local backbones. Finally, fixed-backbone ablations assess the effects of removing L2S, S2L, and components of the L2S--S2L control loop, with runtime-event analysis examining recovery after execution problems.

\subsection{Experimental Setup}

\paragraph{Benchmarks.}
We evaluate LocalLSTC on OSWorld-Verified with the Google Drive tasks included \citep{xie2024osworld} as the primary benchmark and WindowsAgentArena \citep{bonatti2024windowsagentarena} as a secondary benchmark for cross-environment evaluation. OSWorld contains 369 Linux tasks, while WindowsAgentArena contains 154 Windows tasks. Most analyses are conducted on OSWorld, with WindowsAgentArena used to assess transfer to a distinct operating environment.

\paragraph{Implementation.}
OSWorld and WindowsAgentArena run in benchmark-defined Ubuntu and Windows virtual machines, respectively, hosted in headless Docker containers. All experiments use screenshots at a resolution of $1280 \times 720$. Local inference runs on a workstation with an Intel Core Ultra 7 265 processor and a 96 GB NVIDIA RTX Pro 6000 Blackwell Workstation Edition GPU. Unless stated otherwise, LocalLSTC uses Qwen3.5-9B \citep{qwen3.5} as the planner backbone and GTA1-7B \citep{yang2025gta1} for visual grounding. Additional experiments evaluate Qwen3.5-4B, Qwen3.6-27B, Qwen3.8-27B, Gemma-4-E4B-it, and Gemma-4-26B-A4B-it \citep{qwen3.5,qwen3.6-27b,qwen3.8,googledeepmind2026gemma4}. Main benchmark results are averaged over two independent runs with the same parameter settings. Unless otherwise stated, other experiments use a single run. Appendix~\ref{app:runtime-parameters} lists runtime parameters, and Appendix~\ref{app:prompts} provides the prompt templates.

\paragraph{Control-failure annotation.}
To characterize the cross-step control failures associated with local-backbone execution, we use GPT-5.6-sol to perform multi-label annotation of OSWorld trajectories from Agent S3, CoAct-1, GTA1, HiSA, and LocalLSTC under the same Qwen3.5-9B planner backbone. The labels \textsc{Stage}, \textsc{Assess}, \textsc{Loop}, \textsc{Recover}, and \textsc{Stop} capture unresolved failures in stage maintenance, progress assessment, repeated no-progress behavior, recovery, and termination. \textsc{Any} indicates that a trajectory contains at least one of these five control failures, \textsc{Other} records a failure outside these categories, and \textsc{None} indicates that no final unresolved failure is present. The complete annotation protocol, including detailed label definitions and boundary rules, is provided in Appendix~\ref{app:control-allocation-guide}. To validate the reliability of the GPT annotations, we stratify OSWorld tasks by domain and randomly sample 37 tasks, yielding 185 trajectories across the five methods. Two PhD-level AI researchers independently annotate the trajectories under blinded method identities using the same protocol.

\paragraph{Metrics.}
We report task success rate (SR) under different step budgets. SR-100, SR-50, and SR-15 impose maximum budgets of 100, 50, and 15 steps. Following OSWorld \citep{xie2024osworld}, one step corresponds to one planner decision. Control-failure comparisons use paired binary outcomes on the same tasks. 

\subsection{Results}

\subsubsection{Backbone Sensitivity and Cross-Step Control Failures}

Existing GUI-agent frameworks are consistently sensitive to backbone strength. We use GPT-5 results from the OSWorld leaderboard\footnote{\url{https://os-world.github.io/}} as frontier references and evaluate Agent S3, GTA1, CoAct-1, and HiSA with Qwen3.5-9B while preserving each framework's native prompts, grounder, action space, tools, search procedure, and 100-step environment budget. As shown in Figure~\ref{fig:core-idea}(b), SR-100 is 23.0--29.6 points lower in the Qwen3.5-9B evaluations, reducing the four-framework average from 60.9\% to 35.2\%. Under the same Qwen3.5-9B backbone, LocalLSTC reaches 47.6\% SR-100, 12.4 points above the four-framework average.

Trajectory annotation further localizes this sensitivity to cross-step control behavior. As shown in Figure~\ref{fig:core-idea}(c), the four baselines have a mean failed-trajectory control-failure incidence of 94.7\% when averaged equally by method. Under the same method-level averaging, \textsc{Stop}, \textsc{Recover}, and \textsc{Assess} are the three most frequent categories, occurring in 78.6\%, 42.2\%, and 34.4\% of failed trajectories, respectively.

Human verification supports the reliability of the aggregate annotation. The two annotators agree on 87.6\% of \textsc{Any} labels, with Cohen's $\kappa=0.751$. Against the two annotators, GPT-5.6-sol achieves \textsc{Any} F1 scores of 84.1\% and 84.9\%, with $\kappa=0.632$ and $0.637$, respectively. Appendix~\ref{app:annotation-human-verification} reports the complete validation results.

Across all OSWorld tasks, LocalLSTC reduces the incidence of observable cross-step control-failure manifestations, with \textsc{Any} decreasing to 41.7\% from a 63.7\% average across the four baselines, a reduction of 22.0 percentage points. LocalLSTC also achieves the highest \textsc{None} incidence at 42.5\%. Appendix~\ref{app:annotation-results} reports the complete label distribution and uncertainty analysis.

\subsubsection{Benchmark Results and Runtime Behavior}

\begin{table*}[t]
\centering
\caption{Success rates on OSWorld and WindowsAgentArena under step budgets of 100, 50, and 15. LocalLSTC results report mean$\pm$SD over two runs. Other methods use their reported results. Bold and underlining mark the highest and second-highest values, and ``-'' indicates an unreported result.}
\label{tab:results}
\setlength{\tabcolsep}{4.0pt}
\renewcommand{\arraystretch}{0.94}
\resizebox{0.99\textwidth}{!}{
\begin{tabular}{llrrrrrr}
\TableTopRule
\theadrow
\multirow{2}{*}{\textbf{{Access}}} & \multirow{2}{*}{\textbf{{Method / Backbone}}}
& \multicolumn{3}{c}{\textbf{{OSWorld}}} & \multicolumn{3}{c}{\textbf{{WindowsAgentArena}}} \\
\cmidrule(lr){3-5}\cmidrule(lr){6-8}
\subheadrow
& & \metricup{SR-100} & \metricup{SR-50} & \metricup{SR-15}
& \metricup{SR-100} & \metricup{SR-50} & \metricup{SR-15} \\
\midrule
\multirow{20}{*}{\sectiontag{API}}
& o3~\citep{openai2025o3o4mini}                                                       &          23.0 &          17.2 &           9.1 &             - &             - &             - \\
& Agent S / GPT-4o~\citep{agashe2024agents}                                           &             - &             - &          20.6 &             - &             - &          18.2 \\
& Navi / GPT-4V-1106~\citep{bonatti2024windowsagentarena}                              &             - &             - &             - &             - &             - &          19.5 \\
& computer-use-preview~\citep{openai2025computerusingagent}                           &          30.5 &          31.3 &          26.0 &             - &             - &             - \\
& Claude 3.7 Sonnet~\citep{anthropic2025claude37}                                     &          35.6 &          35.8 &          27.1 &             - &             - &             - \\
& Claude 4 Sonnet~\citep{anthropic2025claude4}                                        &          41.4 &          43.9 &          31.2 &             - &             - &             - \\
& Agent S2 / Gemini 2.5 Pro~\citep{agashe2025agents2}                                 &             - &          45.8 &          34.6 &             - &          41.4 &             - \\
& GTA1-32B / o3~\citep{yang2025gta1}                                                  &          53.1 &          48.6 &             - &          51.2 &             - &             - \\
& UI-TARS-2-2509~\citep{wang2025uitars2}                                              &          53.1 &             - &             - &             - &          50.6 &             - \\
& CoAct-1 / CUA 4o + o3 + o4-mini~\citep{song2025coact}                               &          59.9 &          56.4 &          39.8 &          52.5 &          43.5 &          21.4 \\
& Agentic-Lybic-Maestro~\citep{guo2025agenticlybic}                                   &          61.9 &          56.9 &             - &             - &             - &             - \\
& Claude Sonnet 4.5~\citep{anthropic2025sonnet45}                                     &          62.9 &          58.1 &          42.9 &             - &             - &             - \\
& Agent S3 / GPT-5~\citep{gonzalez2025scaling}                                        &          63.5 &             - &             - &          50.2 &          49.0 &             - \\
& UiPath Screen Agent / Opus 4.5~\citep{cristescu2025uicube}                          &          67.1 &          64.4 &             - &             - &             - &             - \\
& \makecell[l]{Agent S3 / GPT-5 bBoN (N=10)~\\\citep{gonzalez2025scaling}}             &          69.9 &             - &             - &          56.6 &          54.1 &             - \\
& VLAA-GUI / Gemini 3 Flash~\citep{han2026vlaagui}                                    &             - &             - &             - &          \second{61.0} &          60.4 &             - \\
& OS-Symphony / GPT-5~\citep{yang2026ossymphony}                                      &             - & \second{65.8} &             - &             - & \second{63.5} &             - \\
& Claude Sonnet 4.6~\citep{anthropic2025sonnet46}                                     &          72.1 &             - &             - &             - &             - &             - \\
& \makecell[l]{Agent S3 / Opus 4.5 + GPT-5 bBoN (N=10)~\\\citep{gonzalez2025scaling}} &          72.6 &             - &             - &             - &             - &             - \\
& VLAA-GUI / Opus 4.5~\citep{han2026vlaagui}                                          & \textbf{76.3} &             - &             - &             - &             - &             - \\
\TableGroupRule
\multirow{18}{*}{\sectiontag{Local}}
& Qwen2.5-VL-7B~\citep{bai2025qwen25vl}                                               &             - &             - &             - &             - &           3.4 &           3.4 \\
& Qwen2.5-VL-72B~\citep{bai2025qwen25vl}                                              &           5.0 &             - &           4.4 &             - &           9.7 &             - \\
& Kimi-VL-A3B~\citep{du2025kimivl}                                                    &          10.3 &             - &           9.7 &             - &             - &          10.4 \\
& InternVL3.5-8B~\citep{wang2025internvl35}                                           &             - &             - &             - &             - &          10.5 &             - \\
& InternVL3.5-38B~\citep{wang2025internvl35}                                          &             - &             - &             - &             - &          14.5 &             - \\
& OpenCUA-Qwen2-7B~\citep{wang2025opencua}                                            &          23.1 &          20.6 &          19.9 &             - &             - &             - \\
& OpenCUA-7B~\citep{wang2025opencua}                                                  &          26.6 &          28.2 &          24.3 &             - &             - &             - \\
& UI-TARS-72B-DPO~\citep{qin2025uitars}                                               &          27.1 &          25.8 &          24.0 &             - &          17.9 &             - \\
& UI-TARS-1.5-7B~\citep{qin2025uitars}                                                &          27.4 &          27.3 &          24.5 &             - &          15.9 &             - \\
& OpenCUA-32B~\citep{wang2025opencua}                                                 &          34.8 &          34.1 &          29.7 &             - &             - &             - \\
& DeepMiner-Mano-7B~\citep{fu2025mano}                                                &          40.2 &             - &             - &             - &             - &             - \\
& OpenCUA-72B-Preview~\citep{wang2025opencua}                                         &          45.0 &          44.9 &          39.0 &             - &             - &             - \\
& OS-Symphony / Qwen3-VL-32B~\citep{yang2026ossymphony}                               &             - &             - &             - &             - &          45.3 &             - \\
& EvoCUA-8B-20260105~\citep{xue2026evocua}                                            &             - &          46.1 &             - &             - &             - &             - \\
& AutoGLM-OS-9B~\citep{lai2025computerrl}                                             &             - &          47.3 &          46.3 &             - &             - &             - \\
& DeepMiner-Mano-72B~\citep{fu2025mano}                                               &          53.9 &             - &             - &             - &             - &             - \\
& GUI-Owl-1.5-32B~\citep{xu2026mobileagent}                                           &             - &          55.4 &             - &             - &             - &             - \\
& EvoCUA-20260105~\citep{xue2026evocua}                                               &             - &          56.7 &             - &             - &             - &             - \\
\TableGroupRule
\highlightrow
& LocalLSTC / Gemma-4-E4B-it                                                & 28.1{\scriptsize $\pm$2.1} & 26.9{\scriptsize $\pm$2.7} & 20.3{\scriptsize $\pm$0.6} & 35.9{\scriptsize $\pm$4.6} & 33.7{\scriptsize $\pm$1.6} & 20.7{\scriptsize $\pm$4.6} \\
\highlightrow
& LocalLSTC / Qwen3.5-4B                                                    & 37.3{\scriptsize $\pm$4.8} & 35.5{\scriptsize $\pm$3.3} & 24.7{\scriptsize $\pm$1.2} & 35.9{\scriptsize $\pm$1.6} & 31.7{\scriptsize $\pm$1.3} & 22.5{\scriptsize $\pm$2.0} \\
\highlightrow
& LocalLSTC / Qwen3.5-9B                                                    & 47.6{\scriptsize $\pm$2.7} & 44.5{\scriptsize $\pm$3.4} & 34.8{\scriptsize $\pm$2.1} & 43.1{\scriptsize $\pm$0.5} & 39.2{\scriptsize $\pm$1.3} & 28.8{\scriptsize $\pm$4.1} \\
\highlightrow
& LocalLSTC / Gemma-4-26B-A4B-it                                            & 60.3{\scriptsize $\pm$0.2} & 54.7{\scriptsize $\pm$2.4} & 42.2{\scriptsize $\pm$1.4} & 53.3{\scriptsize $\pm$1.1} & 44.6{\scriptsize $\pm$4.6} & 35.9{\scriptsize $\pm$1.6} \\
\highlightrow
& LocalLSTC / Qwen3.6-27B                                                   & 64.7{\scriptsize $\pm$0.6} & 62.7{\scriptsize $\pm$0.4} & \second{51.4{\scriptsize $\pm$1.9}} & 59.8{\scriptsize $\pm$1.6} & 59.8{\scriptsize $\pm$1.6} & \second{45.7{\scriptsize $\pm$0.0}} \\
\highlightrow
\multirow{-6}{*}{\ourscell{Ours}}
& LocalLSTC / Qwen3.8-27B                                                   & \second{73.7{\scriptsize $\pm$1.8}} & \textbf{72.3{\scriptsize $\pm$2.0}} & \textbf{61.1{\scriptsize $\pm$4.5}} & \textbf{68.5{\scriptsize $\pm$1.6}} & \textbf{67.9{\scriptsize $\pm$0.6}} & \textbf{57.5{\scriptsize $\pm$4.9}} \\
\TableBottomRule
\end{tabular}
}
\end{table*}

Table~\ref{tab:results} summarizes the benchmark results across step budgets. On OSWorld, LocalLSTC with Qwen3.8-27B reaches $73.7\pm1.8\%$ SR-100, $72.3\pm2.0\%$ SR-50, and $61.1\pm4.5\%$ SR-15, achieving the strongest local results at all three step budgets. These results improve the previous strongest local scores by 19.8, 15.6, and 14.8 points, respectively. The advantage remains consistent across different step budgets, suggesting that LocalLSTC maintains stable control across varying interaction horizons and is robust to long-horizon error accumulation.

On WindowsAgentArena, the same configuration reaches $68.5\pm1.6\%$ SR-100, $67.9\pm0.6\%$ SR-50, and $57.5\pm4.9\%$ SR-15, establishing a new state of the art at all three step budgets. Its mean SR-50 improves the previous strongest local result by 22.6 points and exceeds the strongest API result by 4.4 points. The strong results across Ubuntu and Windows indicate that the control architecture generalizes across different operating environments without benchmark-specific training.

LocalLSTC also benefits consistently from stronger local reasoning backbones. Across the evaluated Qwen backbones and within the Gemma-4 family, stronger models improve SR at every evaluated step budget on both benchmarks. This scaling pattern indicates that stronger temporal control and stronger reasoning backbones provide complementary performance gains.

\captionsetup[table]{font=scriptsize}
\begin{table*}[t!]
\centering
\caption{Domain-level runtime behavior of LocalLSTC with Qwen3.5-9B on OSWorld, averaged over two runs. Execution-channel columns report action shares, with the remainder reserved for control actions. Pass, Fail, and Bypass are mutually exclusive trajectory categories based on Final Verification history. Bypass takes precedence for consecutive-stall or maximum-step cutoffs without an explicit \texttt{INFEASIBLE} action. In all other cases, a trajectory is classified as Pass if it has no rejected Final Verification and as Fail if it has at least one, even when a later verification succeeds. An explicit \texttt{INFEASIBLE} action counts as an accepted Final Verification outcome, with earlier rejections still counted.}
\label{tab:domain_sr}
\setlength{\tabcolsep}{3.0pt}
\renewcommand{\arraystretch}{0.98}
\resizebox{0.95\textwidth}{!}{
\begin{tabular}{lrrrrrrrrrrrrr}
\TableTopRule
\theadrow
\multicolumn{3}{c}{\textbf{{Summary}}} & \multicolumn{4}{c}{\textbf{{Commitment}}} & \multicolumn{3}{c}{\textbf{{Execution Channels}}} & \multicolumn{4}{c}{\textbf{{Evidence Control}}} \\
\cmidrule(lr){1-3}\cmidrule(lr){4-7}\cmidrule(lr){8-10}\cmidrule(lr){11-14}
\subheadrow
\textbf{Domain} & \textbf{Tasks} & \makecell{\metricup{SR} \\ \textbf{(\%)}} & \textbf{steps} & \makecell{\textbf{Subgoals} \\ \textbf{/ Task}} & \makecell{\textbf{Actions} \\ \textbf{/ Subgoal}} & \makecell{\textbf{Actions} \\ \textbf{/ step}} & \makecell{\textbf{GUI} \\ \textbf{(\%)}} & \makecell{\textbf{Bash} \\ \textbf{(\%)}} & \makecell{\textbf{API} \\ \textbf{(\%)}} & \makecell{\textbf{Context} \\ \textbf{Refines}} & \makecell{\textbf{Pass} \\ \textbf{(\%)}} & \makecell{\textbf{Fail} \\ \textbf{(\%)}} & \makecell{\textbf{Bypass} \\ \textbf{(\%)}} \\
\midrule
Chrome & 46 & 59.7 & 20.4 & 7.8 & 2.9 & 1.1 & 86.8 & 1.3 & 2.7 & 2.0 & 64.1 & 13.0 & 22.8 \\
GIMP & 26 & 78.8 & 14.3 & 5.1 & 3.9 & 1.4 & 86.3 & 7.4 & 0.0 & 1.1 & 75.0 & 5.8 & 19.2 \\
LibreOffice Calc & 47 & 28.7 & 30.1 & 11.3 & 4.2 & 1.6 & 75.5 & 5.0 & 13.3 & 3.2 & 42.6 & 20.2 & 37.2 \\
LibreOffice Impress & 47 & 33.4 & 24.7 & 8.1 & 4.7 & 1.6 & 81.0 & 1.6 & 9.5 & 2.2 & 41.5 & 19.1 & 39.4 \\
LibreOffice Writer & 23 & 39.1 & 21.7 & 8.0 & 4.2 & 1.6 & 83.4 & 3.8 & 4.7 & 2.1 & 45.7 & 19.6 & 34.8 \\
Multi-apps & 101 & 32.2 & 36.1 & 14.1 & 3.2 & 1.3 & 65.6 & 25.1 & 2.4 & 3.9 & 55.4 & 16.8 & 27.7 \\
OS & 24 & 83.3 & 17.8 & 6.1 & 3.2 & 1.1 & 38.3 & 54.0 & 0.0 & 1.6 & 77.1 & 14.6 & 8.3 \\
Thunderbird & 15 & 73.3 & 24.9 & 9.5 & 3.3 & 1.3 & 85.7 & 9.6 & 0.0 & 2.7 & 73.3 & 13.3 & 13.3 \\
VLC & 17 & 58.7 & 24.3 & 10.1 & 2.7 & 1.1 & 68.5 & 15.7 & 4.2 & 2.6 & 61.8 & 17.6 & 20.6 \\
VS Code & 23 & 69.6 & 16.4 & 7.5 & 2.8 & 1.3 & 68.2 & 20.6 & 0.7 & 1.6 & 50.0 & 23.9 & 26.1 \\
\midrule
\summaryrow
\keylabel{All} & 369 & 47.6 & 26.1 & 9.9 & 3.5 & 1.3 & 73.1 & 14.3 & 5.2 & 2.7 & 56.0 & 16.7 & 27.4 \\
\TableBottomRule
\end{tabular}
}
\end{table*}

Table~\ref{tab:domain_sr} summarizes LocalLSTC's runtime behavior across OSWorld domains with Qwen3.5-9B. Averaged over two runs, each task uses 26.1 planner steps and 9.9 subgoal segments, with 3.5 actions per subgoal and 1.3 actions per planner step. GUI actions account for 73.1\% of execution, Bash accounts for 14.3\%, and software APIs account for 5.2\%. The execution mix varies substantially across domains. Bash reaches 54.0\% in OS tasks, while GUI interaction dominates Chrome at 86.8\% and GIMP at 86.3\%. LocalLSTC performs 2.7 context refinements per task. Pass, Fail, and Bypass account for 56.0\%, 16.7\%, and 27.4\%, respectively. WindowsAgentArena shows similar runtime patterns, with full statistics reported in Appendix~\ref{app:waa-domain-scores}.

\subsubsection{Ablations and Mechanism Analysis}

We examine which mechanisms account for LocalLSTC's gains under a fixed Qwen3.5-9B backbone. Table~\ref{tab:component-ablation} reports grouped ablations of L2S and S2L together with individual component ablations of the L2S--S2L control loop. Appendix~\ref{app:interface-ablation-definitions} specifies the exact settings for all grouped and component ablations.

\begin{table}[t]
\centering
\caption{Grouped-mechanism and component ablations on OSWorld with Qwen3.5-9B. Resource metrics report per-task averages. Bold marks the highest SR or lowest resource value in each column.}
\label{tab:component-ablation}
\setlength{\tabcolsep}{3.5pt}
\renewcommand{\arraystretch}{0.98}
\resizebox{\linewidth}{!}{
\begin{tabular}{llrrrrrr}
\TableTopRule
\theadrow
\textbf{Category} & \textbf{Method} & \metricup{SR-100} & \metricdown{Tokens (k)} & \metricdown{Prompt (k)} & \metricdown{Completion (k)} & \metricdown{Steps} & \metricdown{Time (s)} \\
\midrule
\highlightrow
\ourscell{Full} & \keylabel{LocalLSTC} & \best{47.6} & 252.2 & 237.6 & 14.6 & 26.1 & 390.4 \\
\midrule
\multirow{2}{*}{\sectiontag{Boundary}} & w/o L2S & 36.4 & \best{192.9} & \best{179.1} & 13.8 & 31.4 & 382.2 \\
& w/o S2L & 41.5 & 246.5 & 229.1 & 17.4 & 27.1 & 472.7 \\
\midrule
\multirow{7}{*}{\sectiontag{Components}}
& w/o persistent-subgoal interface & 38.0 & 276.6 & 257.3 & 19.3 & 29.6 & 554.4 \\
& w/o candidate proposals & 40.7 & 273.0 & 259.1 & 13.9 & 31.2 & 412.5 \\
& w/o multi-action list & 43.5 & 260.7 & 243.0 & 17.8 & 28.8 & 445.5 \\ 
& w/o Step Abstraction & 43.0 & 276.2 & 261.1 & 15.1 & 31.9 & 428.7 \\
& w/o state-conditioned routing & 43.9 & 193.5 & 180.1 & 13.4 & \best{22.8} & \best{365.9} \\
& w/o stall/loop handling & 45.1 & 285.0 & 266.9 & 18.1 & 29.8 & 517.2 \\
& w/o Final Verification & 44.7 & 227.8 & 214.7 & \best{13.1} & 24.1 & 429.0 \\
\TableBottomRule
\end{tabular}
}
\end{table}

Removing L2S lowers SR-100 from 47.6\% to 36.4\%, while removing S2L lowers it to 41.5\%, showing that performance degrades when either grouped mechanism is removed under the same Qwen3.5-9B backbone. Among individual components, removing the persistent-subgoal interface produces the largest reduction of 9.6 points. Removing candidate proposals, Step Abstraction, the multi-action list, state-conditioned routing, Final Verification, and stall/loop handling reduces SR by 6.9, 4.6, 4.1, 3.7, 2.9, and 2.5 points, respectively. These reductions span both commitment formation and cross-step evidence control, supporting contributions from both temporal scopes.

Several components improve both success and interaction efficiency. Removing Step Abstraction increases token use from 252.2k to 276.2k, planner steps from 26.1 to 31.9, and runtime from 390.4 to 428.7 seconds, while removing the persistent-subgoal interface raises token use to 276.6k and runtime to 554.4 seconds. Without candidate proposals, token use and planner steps increase to 273.0k and 31.2. These results suggest that stronger control decisions offset their inference cost by reducing downstream interaction.

\begin{table*}[t]
\centering
\caption{Runtime-event diagnostics for Full, \texttt{w/o L2S}, and \texttt{w/o S2L} on OSWorld. Rate reports the fraction of tasks that enter each event at least once, and Score reports mean evaluator score conditional on event entry. Bold marks the column maximum, and ``--'' indicates an event unavailable after removing the corresponding mechanism.}
\label{tab:trajectory-control-diagnostics}
\footnotesize
\setlength{\tabcolsep}{3.5pt}
\renewcommand{\arraystretch}{0.97}

\begin{tabular}{@{}lrrrrrr@{}}
\TableTopRule
\theadrow
& \multicolumn{2}{c}{\textbf{{Stall / repetition}}}
& \multicolumn{2}{c}{\textbf{{Recovery entry}}}
& \multicolumn{2}{c}{\textbf{{Rejected termination}}} \\
\cmidrule(lr){2-3}\cmidrule(lr){4-5}\cmidrule(lr){6-7}
\subheadrow
\textbf{Method}
& \textbf{Rate (\%)}
& \metricup{Score (\%)}
& \textbf{Rate (\%)}
& \metricup{Score (\%)}
& \textbf{Rate (\%)}
& \metricup{Score (\%)} \\
\midrule

\highlightrow
\ourscell{Full} & 4.5 & \best{34.3} & \best{60.0} & \best{34.0} & \best{32.1} & \best{38.9} \\
\texttt{w/o L2S} & 4.3 & 18.8 & 56.5 & 20.9 & 19.3 & 23.1 \\
\texttt{w/o S2L} & \best{4.6} & 11.8 & -- & -- & -- & -- \\

\TableBottomRule
\end{tabular}
\end{table*}
\captionsetup[table]{font=small}

Table~\ref{tab:trajectory-control-diagnostics} separates event frequency from post-event behavior. Stall or repetition incidence is similar across Full, \texttt{w/o L2S}, and \texttt{w/o S2L} at 4.5\%, 4.3\%, and 4.6\%. Conditional on such events, Full reaches a mean evaluator score of 34.3\%, compared with 18.8\% for \texttt{w/o L2S} and 11.8\% for \texttt{w/o S2L}. These descriptive differences are consistent with improved recovery.

The pattern extends to recovery and termination. Recovery entry and rejected termination are S2L-defined events and are therefore unavailable for \texttt{w/o S2L}. Among the remaining variants, Full reaches 34.0\% after recovery entry compared with 20.9\% for \texttt{w/o L2S}, while the scores after rejected termination are 38.9\% and 23.1\%. These event-conditioned results provide mechanism-specific diagnostics consistent with the grouped ablations, with representative trajectory-level examples provided in Appendix~\ref{app:examples}.

\FloatBarrier
\section{Conclusion}
\looseness=-1

LocalLSTC addresses the cross-step control burden in locally deployed GUI agents by organizing persistent state and short-term execution commitments. Long-to-Short Planning forms each commitment from persistent state, while Short-to-Long Control integrates execution outcomes back for progress assessment, recovery, and termination. LocalLSTC improves task success and reduces observable cross-step control-failure manifestations, with ablations supporting both mechanisms. With Qwen3.8-27B, it reaches 73.7\% SR-100 on OSWorld and 68.5\% on WindowsAgentArena, achieving the strongest local OSWorld result and a new state of the art on WindowsAgentArena. These results show that temporal control organization complements backbone scaling.


\subsection*{Reproducibility Statement}
We provide documentation and artifacts to support reproduction of both the LocalLSTC framework and the reported experiments. Sections~\ref{sec:method} and~\ref{sec:experiments} describe the framework and experimental protocol. Appendix~\ref{app:control-allocation-guide} details the control-failure annotation and human-verification procedures. Appendix~\ref{app:implementation-details} documents implementation details, while Appendix~\ref{app:prompts} provides the prompt templates. Code is released through an anonymous repository to support reproducibility.

\subsection*{AI Use Statement}
GPT-5.6-sol was used for post-hoc trajectory annotation, language polishing, and \LaTeX{} revision. OpenAI Codex was used to retrieve and identify relevant literature and to draft the Related Work section. The authors independently reviewed the AI-generated content and manually verified all cited works and the claims attributed to them against the original sources. All AI-assisted analyses and revisions were reviewed by the authors, who take full responsibility for the accuracy and content of the final manuscript.

\bibliography{main}
\bibliographystyle{iclr2027_conference}

\appendix
\raggedbottom
\section{Implementation Details}
\label{app:implementation-details}

This section specifies how LocalLSTC instantiates the control flow in Section~\ref{sec:method}. Each planner call constructs a Short-Term Execution Commitment from the current observation and Long-Term Control State. The runtime validates and executes the commitment, converts the resulting observations into subgoal-aligned evidence, and uses this evidence to update the persistent control state for the next interaction.

\subsection{Planner Context and Interface}
\label{app:planner-context-interface}

Each planner call conditions on the task instruction, current screenshot, active subgoal, compact evidence history, runtime feedback, available action channels, and domain-specific operational guidance. The planner receives the GUI interaction instructions, platform-specific Bash instructions, and software API schemas available for the relevant applications in the same call. This unified context allows each commitment to select or combine channels based on the current state.

Planner context combines refined history with recent abstracted steps so that compressed cross-step evidence remains available together with recent execution detail. Runtime feedback records loop detections, no-change outcomes, tool failures, and rejected termination. Application-specific skills and software API schemas are initialized from the task-level \texttt{related\_apps} set. For tasks involving multiple applications, a recognized foreground application focuses the active set on that application and the operating-system domain when applicable. The full related set remains active when foreground detection does not identify a matching application.

The planner returns a JSON object containing \texttt{thought}, \texttt{subgoal}, and \texttt{actions}. The \texttt{subgoal} field specifies the execution stage of the current commitment. The reserved value \texttt{continue} retains the active subgoal, while another value proposes a transition to a new subgoal. The runtime canonicalizes the field by stripping surrounding whitespace and collapsing repeated whitespace, then compares the resulting strings exactly. The nonempty \texttt{actions} list specifies the ordered actions executed under the commitment.

Executable actions support GUI interaction, software API calls, and Bash execution. GUI interactions are represented as executable \texttt{pyautogui} code strings. Software API calls use schema-exposed \texttt{Class.method(...)} call strings. The application-specific API schemas and registered execution handlers are adopted directly from ComputerRL \citep{lai2025computerrl}. LocalLSTC changes how these calls are selected and integrated into temporal control. Bash execution uses \texttt{bash\_execution} strings, which the runtime interprets as Bash commands or Python snippets according to the active platform skill. A commitment can combine actions from these channels when a subgoal requires coordinated interaction across interfaces.

The action space also includes the reserved control actions \texttt{WAIT}, \texttt{TERMINATE}, and \texttt{INFEASIBLE}. \texttt{WAIT} retains the active subgoal while the runtime observes for an environment update or timeout. \texttt{TERMINATE} requests Final Verification after preceding executable actions complete. \texttt{INFEASIBLE} records an infeasibility judgment. Terminal actions are placed at the end of a commitment, and validation rejects a commitment that contains both \texttt{TERMINATE} and \texttt{INFEASIBLE}.

The runtime recognizes reserved control actions directly and identifies GUI interaction and software API calls from their registered prefixes. Other executable strings enter the Bash channel, where the active platform skill determines their interpretation. Actions are dispatched in list order, and an execution failure stops the remaining actions in the commitment. Candidate generation and selection occur within the same planner completion. The planner autoregressively emits one to three candidate proposals, followed by a \texttt{Best} index and selection rationale, and finally the executable JSON object. No additional sampling, planner invocation, or external judge is used. Only the final JSON object is parsed for execution.

\subsection{Validation and Canonicalization}
\label{app:validation-details}

Planner outputs are canonicalized before execution through list-level, channel-level, and schema-level validation. List validation enforces terminal-action placement and rejects commitments containing both \texttt{TERMINATE} and \texttt{INFEASIBLE}. Channel canonicalization removes wrapper artifacts from \texttt{bash\_execution} input, converts GUI scripts into executable \texttt{pyautogui} code, and constructs stable fingerprints used by loop control. Software API validation restricts each action to a single callable expression defined by the exposed schema and implemented by a registered handler. Validated parameter literals are converted to runtime values before execution.

The runtime first applies local JSON repair when strict JSON parsing fails. Remaining parsing or validation failures are returned to the planner together with the original response and the corresponding error message. The repair interface allows up to three attempts for each planner call before the configured runtime failure handling applies.

\subsection{Step Abstraction}
\label{app:step-abstraction-details}

Step Abstraction evaluates each execution outcome against the intended effect of the current commitment and converts channel-specific observations into a common subgoal-aligned evidence representation. Each evidence record retains the executed action, observed result, task-relevant progress, execution exceptions, and relevant environment changes.

For GUI interaction, planner-generated \texttt{pyautogui} code is parsed with an abstract syntax tree (AST) and divided into observation-aligned execution chunks. For each coordinate-dependent statement, the runtime extracts a referring expression from contiguous comments immediately preceding the statement or from its trailing \texttt{\#} comment and passes it as the GTA1-7B target description. If no comment is available, a commitment containing one grounding operation falls back to the planner's \texttt{thought}. A multi-grounding commitment uses a generic description formed from the operation index and current sub-action, while a single operation lacking either source uses a screenshot-conditioned default target description. Planner-provided coordinates are treated as proposals and are normally replaced through GTA1-7B grounding. Drag endpoints are grounded separately, and scrolling may replace or insert a coordinate identifying the scroll region. Non-spatial operations such as \texttt{write}, \texttt{press}, and \texttt{hotkey} bypass visual grounding. After each chunk, the runtime acquires a stable screenshot and compares it with the pre-action screenshot to determine whether the interface changed. The stable frame becomes the observation for the next chunk.

The runtime measures screenshot stability across consecutive grayscale captures using the normalized area of the bounding box enclosing changed pixels. Captures below the configured change threshold extend the stable sequence, while larger changes restart it. Initial, post-action, and pre-evaluation observations apply the stage-specific stability requirements and timeouts summarized in Table~\ref{tab:runtime-parameters}. When stabilization times out, the runtime retains the latest available frame and records a warning. Screenshot acquisition failure yields an unavailable observation. Planner observations and Final Verification use screenshot-availability polling.

To localize visual change, the runtime computes the pixel-wise absolute difference between pre-action and post-action screenshots and extracts the bounding box of changed pixels with a 50-pixel margin on each side. When no changed region is detected, Step Abstraction uses the full screenshot. The resulting visual transition is interpreted jointly with the task instruction, active subgoal, executed action, and available recovery feedback.

For \texttt{bash\_execution}, the evidence record retains the executed command, execution status, exit code, textual output, and available readback evidence. Long outputs are compressed while preserving task-relevant identifiers such as paths, filenames, URLs, IDs, table names, column names, and selected values. Software API evidence similarly retains the callable, execution status, return value, helper observation, and error message. GUI-mediated API execution additionally follows the visual observation policy above. All channels enter compact history through the same evidence interface.

\subsection{State and Loop Control}
\label{app:state-control-details}

After Step Abstraction, the runtime assigns a transient routing state using a fixed precedence. \texttt{INFEASIBLE} is handled as a direct runtime failure before the routing-state ladder. A final \texttt{TERMINATE} request maps to \texttt{finish}. Execution failure maps to \texttt{error}. Repeated no-progress evidence maps to \texttt{stall} when the proposed subgoal remains the current subgoal. Otherwise, a changed subgoal maps to \texttt{advance}. Remaining executions map to \texttt{continue}. State assignment occurs before a proposed subgoal transition is committed, keeping the resulting evidence aligned with the execution subgoal that produced it.

The following pseudocode makes the subgoal resolution, routing precedence, and refinement trigger explicit. Here, \texttt{normalize} strips surrounding whitespace and collapses repeated whitespace, and $I_{\min}$ and $I_{\max}$ are the configured minimum and maximum refinement intervals.

\begin{lstlisting}[basicstyle=\small\ttfamily, columns=fullflexible, breaklines=true, frame=single]
raw = normalize(decision.subgoal)
proposed = current_subgoal if raw.lower() == "continue" else raw
steps_since_refinement += 1

if INFEASIBLE was reached:
    status = failure
    return

if execution was suppressed by loop control:
    record(loop_stall)
    consecutive_stall_count += 1
    feedback = "Switch target/tool/approach before retrying"
    if steps_since_refinement >= I_max:
        Refine(accumulated_evidence)
        steps_since_refinement = 0
    if consecutive_stall_count >= 3:
        status = failure
        return
    status = running
    return

executed = outcome.executed_prefix
evidence = Update(evidence, SA(proposed, executed, outcome))

if TERMINATE was reached:              route = finish
elif execution failed:                 route = error
elif proposed == current_subgoal and 3 no-change steps under current_subgoal:
    route = stall
elif proposed != current_subgoal:      route = advance
else:                                  route = continue

if route in {continue, advance}: feedback = None
if route == advance: current_subgoal = proposed
if route == error:   feedback = execution_error
if route == stall:   feedback = no_progress
if route == finish:
    if Verify(evidence):
        status = success
        return
    route = stall
    feedback = rejected_termination

if route == stall: consecutive_stall_count += 1
else:              consecutive_stall_count = 0
if consecutive_stall_count >= 3:
    status = failure
    return

if steps_since_refinement >= I_min and route == advance:
    Refine(accumulated_evidence)
    steps_since_refinement = 0
elif steps_since_refinement >= I_max:
    Refine(accumulated_evidence)
    steps_since_refinement = 0

status = running
\end{lstlisting}

No-change evidence is aggregated once per planner step, so a multi-action commitment contributes at most one no-change observation. Three consecutive no-change steps under the same subgoal produce \texttt{stall}. A proposed transition to a different subgoal takes the \texttt{advance} route before the same-subgoal stall rule applies. Steps containing \texttt{bash\_execution} are excluded from visual no-change aggregation because textual readback may establish progress in the absence of a corresponding interface change. Each \texttt{stall} increments the consecutive-stall counter, while any other route resets it. Three consecutive stalls reach the runtime failure boundary. Runtime feedback is transition-local. Nominal \texttt{continue} and \texttt{advance} routes clear prior feedback, while \texttt{error}, \texttt{stall}, and rejected termination write feedback for the next planner call. The refinement counter increments once per planner step and resets after each refinement.

Loop detection compares normalized action and result fingerprints across interactions and suppresses execution when repetition thresholds are reached. Before GUI/API/Bash dispatch, the runtime performs two checks. First, for a single \texttt{gui\_action}, a third consecutive occurrence of the identical raw \texttt{pyautogui} string records the current step as \texttt{loop\_stall} and skips that GUI execution. Second, repeated action--result pairs suppress the next execution after three repetitions for \texttt{bash\_execution} and software API actions or five repetitions for GUI actions. A suppressed decision still consumes one planner step, increments the consecutive-stall counter, checks the maximum-interval fallback for Context Refinement, and injects recovery feedback that directs the next planner call to change the target, tool, or approach. This suppression occurs before execution. The no-change rule above applies after execution. Three consecutive stall events terminate the task with failure. Because the inserted \texttt{loop\_stall} record interrupts the sequence of identical GUI logs, suppression blocks the current repetition and allows the action to be reconsidered after replanning. Repetition counts use at most one result from each planner step, preventing count inflation from multiple actions within one commitment.

The evidence record combines visual observations, execution logs, textual readback, software API returns, and abstracted steps. Textual readback may come from command output or subsequent file inspection. GUI context remains available across text-based actions, allowing confirmed textual evidence to support subsequent planning and Final Verification together with visual evidence.

\subsection{Context Refinement}
\label{app:context-refinement-details}

LocalLSTC adopts HiSA's context-refinement operation and schedules it through execution-state triggers in place of a fixed period. An \texttt{advance} triggers refinement of accumulated evidence once the minimum refinement interval is reached. The first refinement summarizes the available abstracted history, while later refinements incorporate only steps not already represented in the existing summary. A maximum refinement interval provides a fallback for long trajectories without an earlier trigger.

Subsequent planner calls and Final Verification receive the refined summary together with recent uncompressed steps. This preserves compact cross-step context while retaining local evidence from the most recent interactions.

\subsection{Final Verification}
\label{app:verification-details}

A \texttt{TERMINATE} request invokes Final Verification using the task instruction, active subgoal, compact execution history, initial screenshot, and latest screenshot. When refined context is available, the verifier combines the summary with abstracted steps not yet represented in it. Otherwise, verification uses the full compact history. For office-style tasks modified through \texttt{bash\_execution}, the runtime may reopen the target file before verification so that the current interface provides refreshed visual evidence.

The verifier returns \texttt{PASS} or \texttt{FAIL}, with invalid responses mapped to \texttt{FAIL}. A \texttt{PASS} confirms termination. A \texttt{FAIL} records the rejected termination as a \texttt{stall} and returns execution to planning with recovery feedback that requests targeted verification or completion of the remaining requirement. The subsequent planner call receives this feedback together with a newly acquired current screenshot. Rejected termination follows the same consecutive-stall failure boundary as other stall events.

\subsection{Runtime Parameters}
\label{app:runtime-parameters}

Table~\ref{tab:runtime-parameters} reports the runtime parameters shared across LocalLSTC experiments. 

\begin{table}[t]
\centering
\small
\setlength{\tabcolsep}{3pt}
\caption{Runtime parameters shared across LocalLSTC experiments.}
\label{tab:runtime-parameters}
\begin{tabular}{p{0.58\linewidth}p{0.22\linewidth}}
\TableTopRule
\theadrow
\textbf{Rule or hyperparameter} & \textbf{Value} \\
\midrule
Model-inference temperature & 0 \\
Model-inference \texttt{top\_p} & 0.95 \\
Model-inference \texttt{top\_k} & 20 \\
Model-inference sampling seed & 42 \\
Model-inference thinking token budget & 16384 \\
Maximum candidate proposals per inference step & 3 \\
Minimum context-refinement interval & 5 steps \\
Maximum context-refinement interval & 20 steps \\
\texttt{bash\_execution} timeout & 180 seconds \\
Maximum planner parsing or repair attempts & 3 \\
Post-action observation timeout & 10 seconds \\
Explicit \texttt{WAIT} timeout & 20 seconds \\
Screenshot acquisition timeout & 6 seconds \\
Evaluation retry timeout & 25 seconds \\
Initial / pre-evaluation screenshot stabilization & 5 low-change observations, 30-second timeout \\
Post-action screenshot stability check & 2 stable frames, 10-second timeout \\
Planner-side repeated GUI decision threshold & 3 consecutive planner calls \\
Repeated GUI action--result threshold & 5 repetitions \\
Repeated \texttt{bash\_execution} or software API action--result threshold & 3 repetitions \\
No-change threshold under the same subgoal & 3 consecutive abstracted steps \\
Consecutive stall-to-fail threshold & 3 stalls \\
\TableBottomRule
\end{tabular}
\end{table}

\section{Prompt Templates}
\label{app:prompts}

This appendix reports the core prompt templates used by the full LocalLSTC configuration in our desktop experiments. Each planner call concatenates the global planner prompt, response-format prompt, enabled software-API guidance, GUI guidance, platform-specific Bash/Python guidance, relevant application API schemas, domain skills, and any active recovery guidance. The API, GUI, and platform-execution guidance remains available together on every call, independent of the preceding action. Prompt changes used by the ablation configurations are summarized in Section~\ref{app:ablation-prompt-variants}.

\subsection{Planner System Prompt}
\label{app:planner-prompt}
\begin{systembox}[title=System Prompt: Global Planner]
\begin{lstlisting}[basicstyle=\small\ttfamily, columns=fullflexible, breaklines=true, breakindent=0pt, frame=none, aboveskip=0pt, belowskip=0pt]
You are an expert in GUI interaction, execution-side automation, and software API tools. Always keep the task instruction in mind.

# General Instructions
- GUI, software API, and platform-execution guidance are available together in every planner call. Choose the shortest reliable channel for the current state, and combine channels when the task requires it.
- Reassess the best channel from the current screenshot, execution history, and runtime feedback. Do not continue using a channel merely because the preceding action used it.

# Action Types
## api
Call one software API exposed for one of the current related apps. Use this when the task maps cleanly to an app-specific software operation. Represent it as a bare `Class.method(...)` action string. Use only methods explicitly listed in the current api tools section. Never invent method names. Use Python literals such as `True`, `False`, and `None`, not JSON/JavaScript literals like `true`, `false`, or `null`.

## gui_action
Execute pyautogui code against the visible GUI. Use this when the task should be completed through visible interface interactions. Represent it as a bare `pyautogui...` action string.

## wait
Wait for async operations to complete and observe UI changes. Represent it as the exact action string `WAIT`.

## bash_execution
Execute execution-side automation through the `bash_execution` tool. This is a historical tool name: on Linux, output Bash commands or Python scripts; on Windows, output Python code snippets executed in the Windows guest. Use this when file-side editing, inspection, or automation is more reliable than GUI interaction. Follow the platform-specific skill section for paths, commands, and examples. Represent it as a bare command/code action string.

## termination
Request task completion after the required result has been verified. Represent it as the exact action string `TERMINATE` and explain the completion evidence in `thought`.

## infeasible
Declare that the task is objectively impossible to complete. Represent it as the exact action string `INFEASIBLE`.
- Some tasks may be infeasible by design. If required capabilities, variables, or app features are unavailable, use `INFEASIBLE` instead of `TERMINATE` and explain the blocker in `thought`.
- Treat explicit app, method, source, target, format, variable, and final-state constraints as required. Do not satisfy a similar task through a workaround and claim completion.
- Do not infer missing required parameters; verify them if possible, otherwise use `INFEASIBLE`.
\end{lstlisting}
\end{systembox}

\subsection{Planner Input Payload}
\label{app:planner-input-payload}

The planner receives the current screenshot as multimodal image content together with a text payload that exposes the persistent control state. The schematic logical payload follows the order below; the listing illustrates the fields exposed to the planner rather than a literal static prompt template. Unavailable fields are left empty or marked as none. Application-specific skills and software API schemas are supplied separately as system-side guidance.

\begin{systembox}[title=Schematic Planner Input Payload]
\begin{lstlisting}[basicstyle=\small\ttfamily, columns=fullflexible, breaklines=true, breakindent=0pt, frame=none, aboveskip=0pt, belowskip=0pt]
Task instruction: {task_instruction}
Current subgoal: {current_subgoal}
Refined history: {refined_history}
Recent abstracted steps: {recent_steps}
Runtime feedback: {recovery_hint}
Foreground application: {foreground_app}
Current screenshot: [attached image content]
\end{lstlisting}
\end{systembox}

\subsection{Planner Response Format Prompt}
\label{app:planner-format-prompt}
\begin{systembox}[title=System Prompt: Planner Response Format]
\begin{lstlisting}[basicstyle=\small\ttfamily, columns=fullflexible, breaklines=true, breakindent=0pt, frame=none, aboveskip=0pt, belowskip=0pt]
# Response Format
You may think through up to 3 candidate proposals before the final answer, but the executable answer must be only the final JSON object.
List only the strongest plausible candidates. Do not invent weak filler strategies just to reach 3 items.

Use this exact structure:
Candidates:
1. <one-sentence strategy>
[optional] 2. <one-sentence strategy>
[optional] 3. <one-sentence strategy>

Best:
<candidate number>

Why:
<one-sentence selection reason>

```json
{
    "thought": "Brief reasoning about the chosen strategy and current action. Check prerequisites and verify previous result.",
    "subgoal": "Meaningful phase-level objective for the current stage of work, or 'continue' to keep the current one",
    "actions": [
        "pyautogui.click(695, 514) # click the radio button next to Microsoft Bing",
        "pyautogui.click(663, 807) # click Set as default button"
    ]
}
```

Field guide:
- The final JSON object is the only executable output. Everything before it is optional planning text.
- Output 1 to 3 candidates, not always 3.
- Every candidate must be realistic and worth considering for this exact state.
- If one approach is clearly dominant, output just 1 candidate.
- If you list candidates, keep them short and plain text only. Do not put JSON, code blocks, or extra headings inside candidate text.
- End your response with exactly one final JSON object.
- The final JSON object must contain `thought`, `subgoal`, and `actions`.
- `subgoal`: a phase-level objective for a coherent stage, not an action title or method. Avoid labels like "Open menu", "Click link", "Scroll to inspect", or "Run script"; prefer "Navigate to the target page", "Update the spreadsheet", or "Verify the result".
- Use `continue` while the next action still pursues, verifies, or strengthens evidence for the current subgoal. Start a new subgoal only when the target state changes or the current stage is blocked.
- `actions` must be a non-empty ordered list. The environment changes after each action, so plan them in execution order.
- Prefer one meaningful interaction per action item. Even though some action strings can contain multiple low-level operations, you should usually split sequential interactions into separate `actions` items for clarity and better replanning.
- Each action item should be one of:
  - a bare `pyautogui...` string for GUI actions
  - a bare `Class.method(...)` string for API calls
  - a bare command/code string for `bash_execution`
  - the string `WAIT` for `wait`
  - the string `TERMINATE` only when the task is already verified complete, with the completion evidence explained in `thought`
  - the string `INFEASIBLE` only when the task is objectively impossible, with the blocker reason explained in `thought`
- Use exactly one API call per action item.
- Bash example:
```json
{
  "thought": "Locate the exact workbook before editing it.",
  "subgoal": "Locate the target workbook",
  "actions": [
    "find /home/user -type f -name '*.xlsx' 2>/dev/null | head -20"
  ]
}
```
- Good example:
Candidates:
1. Use a short GUI action sequence to open the search engine menu and set Bing there.
2. Use one direct browser API call if an API exists.

Best:
1

Why:
The relevant setting is visible in the UI and there is no confirmed api call for this browser preference.

```json
{
  "thought": "Open the visible browser setting and choose Bing through the interface.",
  "subgoal": "Set Bing as the default search engine",
  "actions": [
    "pyautogui.click(695, 514) # click the radio button next to Microsoft Bing",
    "pyautogui.click(663, 807) # click Set as default button"
  ]
}
```

- Termination example:
```json
{
  "thought": "The required result is already visible and verified in the current screenshot.",
  "subgoal": "Verify task completion",
  "actions": [
    "TERMINATE"
  ]
}
```

- Infeasible example:
```json
{
  "thought": "The requested result cannot be completed because the required app or capability is unavailable in the current environment.",
  "subgoal": "Confirm the blocker",
  "actions": [
    "INFEASIBLE"
  ]
}
```
\end{lstlisting}
\end{systembox}

\subsection{Response Repair Prompt}
\label{app:repair-prompt}
\begin{systembox}[title=System Prompt: Response Repair]
\begin{lstlisting}[basicstyle=\small\ttfamily, columns=fullflexible, breaklines=true, breakindent=0pt, frame=none, aboveskip=0pt, belowskip=0pt]
Repair the planner response below and preserve the intended plan.

Repair rules:
- Focus on fixing the concrete parse/schema/tool-syntax error shown above.
- If the error points to a specific line, column, or token, fix that exact location.
- Action content must be a bare string, not an object.
- A GUI action must be a valid Python `pyautogui` code string.
- An API action must be exactly one Python call expression such as `GoogleDriveTools.upload_file(...)`.
- A shell action must contain only raw shell/Python code. Remove any leading `bash_execution` wrapper, function syntax, colon, or label.
- Return exactly one repaired response.
- The final output must satisfy this response format exactly:

{response_format_prompt}

Original response:
{response}

Parse traceback:
{error_message}
\end{lstlisting}
\end{systembox}

\subsection{Step Abstraction Prompt}
\label{app:step-abstraction-prompt}
\begin{systembox}[title=System Prompt: Step Abstraction]
\begin{lstlisting}[basicstyle=\small\ttfamily, columns=fullflexible, breaklines=true, breakindent=0pt, frame=none, aboveskip=0pt, belowskip=0pt]
Compare the latest observations and summarize what happened in 1-2 concise sentences.

Task instruction: {task_instruction}
Current subgoal: {current_subgoal}
Action: {action_description}
Runtime feedback: {recovery_hint}

Rules:
- Describe what changed, or say no visible change.
- Mention clear errors if shown.
- Do not judge subgoal status or long-horizon task completion.
- Keep the summary concise and concrete.
- For bash/API output, preserve exact task-relevant identifiers from the output, especially paths, filenames, URLs, IDs, sheet/table/column names, window titles, counts, and selected values.
- Use the task instruction and current subgoal to decide which output details are key; do not replace needed paths or IDs with generic wording.
\end{lstlisting}
\end{systembox}

\subsection{Context Refinement Prompt}
\label{app:context-refinement-prompt}
\begin{systembox}[title=System Prompt: Context Refinement]
\begin{lstlisting}[basicstyle=\small\ttfamily, columns=fullflexible, breaklines=true, breakindent=0pt, frame=none, aboveskip=0pt, belowskip=0pt]
Analyze task execution progress and provide guidance.

You will receive: a task instruction, execution history range and execution history

Instructions:
- Summarize the full execution history into one unified summary covering the entire range
- List what was done in order (successes and failures)
- **IMPORTANT**: Preserve coordinates in click actions (e.g., "click(500,300)") so the summary records the exact prior interactions
- Identify repeated no-progress behavior, task-relevant progress, or stalled execution.
- Provide actionable suggestions for the next step if there are issues

Return a concise summary string in this format:
Steps X~Y: [ordered list of what was done, keeping coordinates]. Suggestion: [actionable advice, or 'Continue' if progressing well]

Examples:
- Steps 1~5: Opened file, tried to edit (failed 3 times with permission error), attempted sudo (failed). Suggestion: Try a different approach - copy file to temp location first.
- Steps 1~5: Clicked Submit button at click(850,620), typed text, clicked Save at click(920,580). Suggestion: Continue - forms being filled correctly.
- Steps 1~10: Previously installed package and ran script (steps 1~5). Then verified output, tested functionality (steps 6~10). Suggestion: Continue - good progress.
- Steps 1~15: Clicked the same button 5 times with no response, tried alternative buttons (failed). Suggestion: Try an alternative method, or use INFEASIBLE if the task is objectively impossible.
\end{lstlisting}
\end{systembox}

\subsection{Final Verification Prompt}
\label{app:final-verification-prompt}
\begin{systembox}[title=System Prompt: Final Verification]
\begin{lstlisting}[basicstyle=\small\ttfamily, columns=fullflexible, breaklines=true, breakindent=0pt, frame=none, aboveskip=0pt, belowskip=0pt]
Decide whether the task is fully completed after the observed workflow.

You will receive:
- the original task
- the current subgoal
- the provided execution history
- the initial screenshot from the start of the task
- the latest screenshot from the end of the task

Return exactly one word:
PASS
or
FAIL

Rules:
- Return only `PASS` or `FAIL`.
- Use `PASS` only if the task requirements appear fully satisfied in the provided execution history and screenshots.
- Some tasks may be infeasible by design. If required capabilities, variables, or app features were unavailable, return `FAIL`.
- If there is uncertainty, return `FAIL`.
- The screenshots are primary evidence for visible state. Use the initial screenshot as baseline context and the latest screenshot as the final state to judge.
- If the task outcome is only obvious by comparing before vs after, explicitly use that comparison before deciding.
- If the latest screenshot does not directly show the requested result, return `FAIL`.
- Use the provided execution history to confirm what was actually modified, saved, read back, or verified. If logs only show that a command or API call ran, but do not confirm the requested final result, return `FAIL`.
- If a file was modified through `bash_execution` while the desktop app may still display stale content, prefer explicit readback/verification evidence from the logs plus the latest screenshot of the reopened app state.
- Fail if the logs reveal formatting mistakes, header corruption, partial coverage, or any mismatch with the task requirements.
- Do not treat an explanation of impossibility, a command launch, or a transient status/toast as completion for a task that asked for an actual GUI, file, or configuration result.
- Require exact evidence for explicit app, method, source, target, format, variable, and final-state constraints; return `FAIL` if the workflow guessed, changed, or bypassed them.
- For multi-target tasks (`all`, `both`, `each`, `respectively`), fail unless every requested target is explicitly covered by the logs.
- For relative-date tasks, fail unless the exact resolved absolute date is explicitly covered.
\end{lstlisting}
\end{systembox}

\subsection{Ablation Prompt Variants}
\label{app:ablation-prompt-variants}

The component ablations preserve the full prompts except at the interface being removed. The configuration without a persistent subgoal replaces the reserved \texttt{continue} value with an explicit phase-level objective on every planner call. The configuration without candidate proposals removes the candidate list, \texttt{Best} index, and selection rationale from the response format. The configuration without a multi-action list replaces \texttt{actions} with a single \texttt{action} field. Removing Step Abstraction bypasses its semantic prompt and records deterministic raw execution evidence. Removing state-conditioned routing, stall and loop handling, or Final Verification changes runtime control without introducing another planner prompt.

The \texttt{w/o L2S} configuration uses a task-level planner interface with \texttt{thought} and one \texttt{action}. It retains the same GUI, Bash/Python, software API, and reserved control-action channels as Full. Its Step Abstraction prompt omits the current subgoal, and its Final Verification prompt omits the current subgoal from the supplied evidence. The \texttt{w/o S2L} configuration retains the full planner commitment interface but uses task-conditioned Step Abstraction without the current subgoal or runtime feedback. It removes state-conditioned routing, routing-dependent recovery feedback, stall and loop handling, and Final Verification, while retaining periodic Context Refinement without state-triggered routing.

\section{Control-Failure Annotation Protocol}
\label{app:control-allocation-guide}

This section presents the annotation protocol used for both GPT-5.6-sol annotation and human verification. Both settings follow the same evidence inputs, label definitions, and decision rules, ensuring consistent assignment of control-failure labels across trajectories.

\FloatBarrier
\begin{annotationbox}{Control-Failure Annotation Guide}

\subsection{Annotation Objective and Setup}
\label{app:annotation-setup}

Annotate unresolved cross-step control failures that remain at trajectory termination. Use the task instruction, complete execution logs, and available evaluator evidence to assess the full trajectory. Problems resolved during subsequent execution are excluded. When evaluator evidence conflicts with the execution record, follow the execution record.

\subsection{Label Definitions}
\label{app:annotation-labels}

The annotation set contains five control-failure labels, \textsc{Stage}, \textsc{Assess}, \textsc{Loop}, \textsc{Recover}, and \textsc{Stop}. Evaluate these labels independently, and record each applicable label at most once per trajectory.

\begin{itemize}[leftmargin=*, nosep]
\item \textsc{Stage} records failure to maintain the task stage or its remaining requirements. It applies when execution moves to an incorrect stage, advances before the current stage is satisfied, returns to an earlier stage without justification, or omits a requirement needed to complete the current stage.

\item \textsc{Assess} records an incorrect assessment of the current execution state or action outcome. It applies when trajectory evidence indicates that an action failed or remained incomplete but the agent treats it as successful, or when a successful action is treated as failed.

\item \textsc{Loop} records repetition of a materially equivalent action after evidence that the previous attempt made no progress, without new information, a relevant change in execution conditions, or an adjustment that could alter the outcome.

\item \textsc{Recover} records an ineffective response to a recognized failure or no-progress condition. It applies when a recovery, rollback, or alternative execution path does not address the observed problem and leaves execution unrecovered.

\item \textsc{Stop} records termination without sufficient evidence that the task has been completed. It applies when required task conditions remain unverified or unsatisfied and execution does not resume.

\item \textsc{Other} records a final unresolved failure that does not satisfy any of the five control-failure definitions.

\item \textsc{None} records a trajectory with no unresolved failure at termination.

\item \textsc{Any} indicates that a trajectory contains at least one of the five control failures.
\end{itemize}

\textsc{Other} and \textsc{None} are mutually exclusive with the five control-failure labels and with each other.

\subsection{Decision Procedure and Boundary Rules}
\label{app:annotation-boundaries}

\begin{enumerate}[leftmargin=*, nosep]
\item Review the complete trajectory and identify failures that remain unresolved at termination.
\item Evaluate \textsc{Stage}, \textsc{Assess}, \textsc{Loop}, \textsc{Recover}, and \textsc{Stop} independently, and record every applicable label.
\item Mark \textsc{Any} if at least one control-failure label applies. If none applies, record \textsc{Other} when an unresolved failure remains, or \textsc{None} when no unresolved failure remains.
\item Assign multiple control-failure labels when the trajectory independently satisfies multiple definitions.
\end{enumerate}

Unless the execution record provides clear contrary evidence, an evaluator score of 0 indicates that a failure is present. If the trajectory does not provide clear evidence for any of the five control-failure labels, record \textsc{Other}. Do not record \textsc{None}.

For \textsc{Loop}, an equivalent retry is labeled only when the previous attempt has explicit no-progress evidence. A retry that introduces new information, changes a relevant execution condition, or makes a corrective adjustment is not labeled as \textsc{Loop}.

For \textsc{Recover}, the failure or no-progress condition must first be recognized. The label applies when the subsequent response does not address that condition.

\textsc{Assess} and \textsc{Stop} may co-occur when the agent incorrectly interprets the execution state and subsequently terminates without sufficient completion evidence. These labels describe observable cross-step control-failure manifestations. Upstream perception, grounding, tool, or environment errors may coexist with a control-failure label when the trajectory independently satisfies its definition.

\end{annotationbox}
\FloatBarrier

\subsection{Human Verification}
\label{app:annotation-human-verification}

We validate the GPT-5.6-sol annotations on a stratified sample of 37 OSWorld tasks shared across all five methods, yielding 185 trajectories. The tasks are sampled without replacement across OSWorld domains using random seed 42.

Two PhD-level AI researchers independently annotate all 185 trajectories using the same label definitions and decision procedure. Each annotator receives the task instruction, complete execution logs, and evaluator evidence, without access to method identity or GPT-5.6-sol labels. The two annotators assign identical complete label sets to 63.2\% of trajectories, with a mean Jaccard overlap of 72.3\%. For \textsc{Any}, they achieve 87.6\% agreement, 88.0\% F1, and Cohen's $\kappa=0.751$. Across the seven original labels, Cohen's $\kappa$ ranges from 0.472 for \textsc{Other} to 0.792 for \textsc{Loop}, with a macro average of 0.656. Table~\ref{tab:annotation-interannotator} reports the complete label-wise inter-annotator agreement.

\begin{table*}[t]
\centering
\small
\setlength{\tabcolsep}{6.2pt}
\renewcommand{\arraystretch}{1.08}
\caption{Inter-annotator agreement on 185 trajectories. A1 and A2 report label prevalence for the two independent human annotators. Agreement, F1, and Cohen's $\kappa$ are computed separately for each label.}
\label{tab:annotation-interannotator}
\begin{tabular}{lrrrrr}
\TableTopRule
\theadrow
\textbf{Label} & \makecell{\textbf{A1 prev.} \\ \textbf{(\%)}} & \makecell{\textbf{A2 prev.} \\ \textbf{(\%)}} & \makecell{\metricup{Agreement} \\ \textbf{(\%)}} & \metricup{F1 (\%)} & \metricup{$\kappa$} \\
\midrule
\textsc{Stage}   & 20.0 & 20.5 & 88.6 & 72.0 & 0.649 \\
\textsc{Assess}  & 22.2 & 30.8 & 84.9 & 71.4 & 0.615 \\
\textsc{Loop}    & 25.4 & 22.2 & 92.4 & 84.1 & 0.792 \\
\textsc{Recover} & 28.1 & 18.4 & 87.0 & 72.1 & 0.641 \\
\textsc{Stop}    & 25.4 & 23.8 & 87.6 & 74.7 & 0.665 \\
\textsc{Other}   & 9.7  & 10.8 & 90.3 & 52.6 & 0.472 \\
\textsc{None}    & 40.0 & 36.2 & 88.6 & 85.1 & 0.760 \\
\summaryrow
\ourscell{\textsc{Any}} & 50.3 & 53.0 & 87.6 & 88.0 & 0.751 \\
\TableBottomRule
\end{tabular}
\end{table*}

We further compare the GPT-5.6-sol annotations with each human annotator on the same 185 trajectories. Table~\ref{tab:annotation-classification-metrics} reports precision, recall, F1, and Cohen's $\kappa$ for each label and for \textsc{Any}, using each annotator independently as the reference. For \textsc{Any}, GPT-5.6-sol achieves 84.1\% F1 with $\kappa=0.632$ against Annotator~1 and 84.9\% F1 with $\kappa=0.637$ against Annotator~2. The corresponding human--human comparison reaches 88.0\% F1 with $\kappa=0.751$.

\begin{table*}[t]
\centering
\small
\setlength{\tabcolsep}{4.8pt}
\renewcommand{\arraystretch}{1.02}
\caption{Agreement between GPT-5.6-sol and the two independent human annotators on the same 185 trajectories. Precision, recall, F1, and Cohen's $\kappa$ are computed separately with each annotator as the reference.}
\label{tab:annotation-classification-metrics}
\begin{tabular}{lrrrrrrrr}
\TableTopRule
\theadrow
& \multicolumn{4}{c}{\textbf{{Annotator 1}}} & \multicolumn{4}{c}{\textbf{{Annotator 2}}} \\
\cmidrule(lr){2-5}\cmidrule(lr){6-9}
\subheadrow
\textbf{Label} & \metricup{P (\%)} & \metricup{R (\%)} & \metricup{F1 (\%)} & \metricup{$\kappa$} & \metricup{P (\%)} & \metricup{R (\%)} & \metricup{F1 (\%)} & \metricup{$\kappa$} \\
\midrule
\textsc{Stage}   & 68.4 & 70.3 & 69.3 & 0.615 & 60.5 & 60.5 & 60.5 & 0.503 \\
\textsc{Assess}  & 62.5 & 73.2 & 67.4 & 0.572 & 62.5 & 52.6 & 57.1 & 0.403 \\
\textsc{Loop}    & 88.4 & 80.9 & 84.4 & 0.795 & 74.4 & 78.0 & 76.2 & 0.692 \\
\textsc{Recover} & 64.1 & 96.2 & 76.9 & 0.652 & 39.7 & 91.2 & 55.4 & 0.400 \\
\textsc{Stop}    & 61.3 & 80.9 & 69.7 & 0.574 & 48.4 & 68.2 & 56.6 & 0.399 \\
\textsc{Other}   & 90.9 & 55.6 & 69.0 & 0.665 & 54.5 & 30.0 & 38.7 & 0.336 \\
\textsc{None}    & 90.6 & 64.9 & 75.6 & 0.634 & 88.7 & 70.1 & 78.3 & 0.681 \\
\summaryrow
\ourscell{\textsc{Any}} & 74.4 & 96.8 & 84.1 & 0.632 & 76.9 & 94.9 & 84.9 & 0.637 \\
\TableBottomRule
\end{tabular}
\end{table*}

\subsection{Annotation Results}
\label{app:annotation-results}

Table~\ref{tab:control-allocation-problems} reports the complete task-level label distribution across all 369 OSWorld tasks under the fixed Qwen3.5-9B planner backbone. LocalLSTC has the lowest \textsc{Any} incidence at 41.7\% and the highest \textsc{None} incidence at 42.5\%, compared with an average \textsc{Any} incidence of 63.7\% across the four baselines.

\begin{table*}[t!]
\centering
\caption{Task-level incidence of control-failure labels on OSWorld under the fixed Qwen3.5-9B planner backbone. Each cell reports the percentage of all 369 tasks assigned the corresponding label.}
\label{tab:control-allocation-problems}
\setlength{\tabcolsep}{3.2pt}
\renewcommand{\arraystretch}{1.10}
\begin{tabular}{lrrrrrrrr}
\TableTopRule
\theadrow
\textbf{Method} & \metricdown{\textsc{Stage}} & \metricdown{\textsc{Assess}} & \metricdown{\textsc{Loop}} & \metricdown{\textsc{Recover}} & \metricdown{\textsc{Stop}} & \metricup{\textsc{None}} & \metricdown{\textsc{Other}} & \metricdown{\textsc{Any}} \\
\midrule
HiSA & 36.3\% & 19.2\% & 14.1\% & 28.2\% & 51.2\% & 31.2\% & 12.7\% & 56.1\% \\
CoAct-1 & 10.3\% & 27.6\% & 27.4\% & 38.2\% & 47.2\% & 28.5\% & 3.8\% & 67.8\% \\
GTA1 & 11.7\% & 23.0\% & 26.3\% & 31.4\% & 49.1\% & 32.2\% & 5.4\% & 62.3\% \\
Agent S3 & 10.8\% & 23.6\% & 11.7\% & 17.3\% & 64.8\% & 28.2\% & 3.3\% & 68.6\% \\
\highlightrow
\ourscell{LocalLSTC} & 18.2\% & 21.4\% & 28.5\% & 33.6\% & 26.6\% & 42.5\% & 15.7\% & 41.7\% \\
\TableBottomRule
\end{tabular}
\end{table*}

\section{WindowsAgentArena Runtime Diagnostics}
\label{app:waa-domain-scores}

Table~\ref{tab:waa-domain-scores} summarizes LocalLSTC's runtime behavior across WindowsAgentArena domains with Qwen3.5-9B and complements the OSWorld analysis in Table~\ref{tab:domain_sr}.

\begin{table}[!ht]
\centering
\caption{Domain-level runtime behavior of LocalLSTC with Qwen3.5-9B on WindowsAgentArena, averaged over two runs. Execution-channel columns report action shares, with the remainder reserved for control actions. Pass, Fail, and Bypass are mutually exclusive trajectory categories based on Final Verification history. Bypass takes precedence for consecutive-stall or maximum-step cutoffs without an explicit \texttt{INFEASIBLE} action. Otherwise, Pass indicates no rejected Final Verification, while Fail indicates at least one, even if a later verification succeeds. An explicit \texttt{INFEASIBLE} action counts as an accepted Final Verification outcome, with earlier rejections still counted.}
\label{tab:waa-domain-scores}
\setlength{\tabcolsep}{4pt}
\renewcommand{\arraystretch}{1.08}
\resizebox{\linewidth}{!}{
\begin{tabular}{lrrrrrrrrrrrrr}
\TableTopRule
\theadrow
\multicolumn{3}{c}{\textbf{{Summary}}} & \multicolumn{4}{c}{\textbf{{Commitment}}} & \multicolumn{3}{c}{\textbf{{Execution Channels}}} & \multicolumn{4}{c}{\textbf{{Evidence Control}}} \\
\cmidrule(lr){1-3}\cmidrule(lr){4-7}\cmidrule(lr){8-10}\cmidrule(lr){11-14}
\subheadrow
\textbf{Domain} & \textbf{Tasks} & \makecell{\metricup{SR} \\ \textbf{(\%)}} & \textbf{steps} & \makecell{\textbf{Subgoals} \\ \textbf{/ Task}} & \makecell{\textbf{Actions} \\ \textbf{/ Subgoal}} & \makecell{\textbf{Actions} \\ \textbf{/ step}} & \makecell{\textbf{GUI} \\ \textbf{(\%)}} & \makecell{\textbf{Bash} \\ \textbf{(\%)}} & \makecell{\textbf{API} \\ \textbf{(\%)}} & \makecell{\textbf{Context} \\ \textbf{Refines}} & \makecell{\textbf{Pass} \\ \textbf{(\%)}} & \makecell{\textbf{Fail} \\ \textbf{(\%)}} & \makecell{\textbf{Bypass} \\ \textbf{(\%)}} \\
\midrule
Chrome & 17 & 34.7 & 24.1 & 8.1 & 3.3 & 1.1 & 84.5 & 0.9 & 5.3 & 2.0 & 35.3 & 23.5 & 41.2 \\
Clock & 4 & 37.5 & 34.5 & 10.5 & 5.5 & 1.7 & 96.0 & 0.0 & 0.0 & 3.4 & 50.0 & 0.0 & 50.0 \\
File Explorer & 19 & 63.2 & 18.5 & 7.8 & 3.0 & 1.3 & 73.4 & 14.4 & 0.0 & 1.6 & 50.0 & 28.9 & 21.1 \\
LibreOffice Calc & 24 & 8.3 & 34.7 & 13.8 & 4.5 & 1.8 & 69.3 & 1.1 & 23.3 & 3.9 & 12.5 & 16.7 & 70.8 \\
LibreOffice Writer & 19 & 7.9 & 24.4 & 11.2 & 3.1 & 1.4 & 79.4 & 0.2 & 7.3 & 2.4 & 18.4 & 7.9 & 73.7 \\
Microsoft Paint & 3 & 83.3 & 22.5 & 10.0 & 3.0 & 1.3 & 88.3 & 4.1 & 0.0 & 2.3 & 50.0 & 33.3 & 16.7 \\
Microsoft Edge & 13 & 50.0 & 19.3 & 8.5 & 2.5 & 1.1 & 88.7 & 1.8 & 0.0 & 1.9 & 57.7 & 19.2 & 23.1 \\
Notepad & 2 & 100.0 & 15.2 & 9.0 & 2.3 & 1.3 & 73.1 & 15.1 & 0.0 & 1.8 & 25.0 & 75.0 & 0.0 \\
Settings & 5 & 100.0 & 6.2 & 4.4 & 1.3 & 0.9 & 85.1 & 0.0 & 0.0 & 0.6 & 100.0 & 0.0 & 0.0 \\
VLC & 21 & 49.6 & 33.1 & 11.5 & 3.5 & 1.2 & 82.9 & 8.6 & 2.8 & 3.4 & 40.5 & 23.8 & 35.7 \\
VS Code & 24 & 62.5 & 21.8 & 8.9 & 3.5 & 1.4 & 88.9 & 4.2 & 0.9 & 2.4 & 62.5 & 10.4 & 27.1 \\
Windows Calculator & 3 & 66.7 & 41.2 & 8.7 & 8.2 & 1.7 & 91.1 & 5.9 & 0.0 & 2.8 & 33.3 & 66.7 & 0.0 \\
\midrule
\summaryrow
\keylabel{All} & 154 & 43.1 & 25.4 & 10.0 & 3.6 & 1.4 & 80.1 & 4.0 & 8.1 & 2.6 & 40.9 & 19.2 & 39.9 \\
\TableBottomRule
\end{tabular}
}
\end{table}

Averaged over two runs, each task uses 25.4 planner steps and 10.0 subgoal segments, with 3.6 actions per subgoal and 1.4 actions per planner step. GUI actions account for 80.1\% of execution, Bash accounts for 4.0\%, and software APIs account for 8.1\%. The execution mix varies across domains. Bash reaches 15.1\% in Notepad and 14.4\% in File Explorer, while software API use reaches 23.3\% in LibreOffice Calc. GUI interaction remains the primary channel in every domain. LocalLSTC performs 2.6 context refinements per task. Pass, Fail, and Bypass account for 40.9\%, 19.2\%, and 39.9\%, respectively.

\section{Ablation Configurations}
\label{app:interface-ablation-definitions}

All ablations use the same OSWorld setup, Qwen3.5-9B planner, GTA1-7B visual grounder, 100-step budget, and inference parameters as Full. Each reported variant changes only the listed configuration flag. All unlisted settings remain unchanged.

\subsection{Grouped L2S--S2L Ablations}

The \texttt{w/o L2S} configuration uses \texttt{--wo\_l2s}. It removes subgoal-conditioned commitment formation and uses task-level single-action planning while retaining all execution channels, task-level post-execution control, and Step Abstraction. The \texttt{w/o S2L} configuration uses \texttt{--wo\_s2l}. It retains the L2S commitment interface but replaces evidence-conditioned post-execution state updates with planner-intent subgoal updates. Table~\ref{tab:legacy-interface-matrix} specifies the resulting interfaces.

\begin{table}[t!]
\centering
\scriptsize
\setlength{\tabcolsep}{4pt}
\renewcommand{\arraystretch}{1.12}
\caption{Grouped L2S and S2L ablation configurations under the fixed Qwen3.5-9B backbone. Full denotes the complete LocalLSTC configuration.}
\label{tab:legacy-interface-matrix}
\resizebox{\textwidth}{!}{
\begin{tabular}{p{0.20\textwidth}>{\columncolor{TableSummary}}p{0.26\textwidth}p{0.26\textwidth}p{0.26\textwidth}}
\TableTopRule
\theadrow
\textbf{Configuration} & \textbf{Full} & \textbf{w/o L2S} & \textbf{w/o S2L} \\
\midrule

Planning context &
Persistent $g_t$, subgoal-aligned evidence, refined history, and recent abstracted steps &
Task-conditioned evidence and history without a persistent subgoal in planner context &
Planner retains a subgoal proposal, while evidence context is task-conditioned \\

Proposal and commitment &
Up to three proposals compared within one inference, followed by an ordered action list &
Single proposal with one action per planner step &
Same as Full \\

Step Abstraction &
Subgoal-conditioned evidence across execution channels &
Retained with task-level conditioning &
Retained with task-level conditioning \\

State update and routing &
Evidence-conditioned \texttt{continue}, \texttt{advance}, \texttt{error}, \texttt{stall}, and \texttt{finish} routing &
Task-level error, stall, and finish handling without subgoal-based stage transitions &
Planner-intent state updates without S2L routing \\

Recovery feedback &
State-conditioned error and stall feedback &
Retained at the task level &
Removed \\

Context Refinement &
\texttt{advance}-triggered refinement with maximum-interval fallback &
Periodic refinement without subgoal advance &
Periodic refinement without state routing \\

Termination &
Final Verification after \texttt{TERMINATE}, with rejection feedback returned to planning &
Task-level Final Verification without subgoal context &
Termination without Final Verification \\

\TableBottomRule
\end{tabular}
}
\end{table}

\subsection{Component Ablations}

Table~\ref{tab:component-ablation-settings} specifies the seven component ablations together with the behavior retained or substituted in each configuration. Each comparison measures the effect of removing the named component together with its dependent runtime behavior. Removing state-conditioned routing also bypasses recovery, stall and loop enforcement, and Final Verification, while removing stall and loop handling also disables error routing and its recovery feedback.

\begin{table}[t!]
\centering
\scriptsize
\setlength{\tabcolsep}{3.5pt}
\renewcommand{\arraystretch}{1.12}
\caption{Component ablation settings under the fixed Qwen3.5-9B backbone.}
\label{tab:component-ablation-settings}
\resizebox{\textwidth}{!}{
\begin{tabular}{p{0.19\textwidth}p{0.36\textwidth}p{0.41\textwidth}}
\TableTopRule
\theadrow
\textbf{Ablation} & \textbf{Removed behavior} & \textbf{Retained or replacement behavior} \\
\midrule

Persistent-subgoal interface &
The planner does not receive the previous subgoal and cannot use \texttt{continue}. &
Each call generates a new subgoal label. Candidate proposals, multi-action commitments, all execution channels, evidence, routing, refinement, and Final Verification remain enabled. \\

Candidate proposals &
The planner omits the candidate list, \texttt{Best} index, and selection rationale. &
The planner directly returns one final JSON decision. Persistent subgoals, ordered action lists, all execution channels, and S2L remain enabled. \\

Multi-action list &
The final decision cannot contain an ordered \texttt{actions} list. &
The response uses one \texttt{action} per planner call. Candidate proposals, persistent subgoals, all execution channels, and S2L remain enabled. \\

Step Abstraction &
The semantic Step Abstraction model call is removed. &
Deterministic raw evidence records the action and observed change for GUI execution. Bash and API evidence additionally records status, exit code, and raw output. Routing, refinement, and Final Verification remain enabled. \\

State-conditioned routing &
Assigned \texttt{continue}, \texttt{advance}, \texttt{error}, \texttt{stall}, and \texttt{finish} states are logged but not applied to runtime control. &
Step Abstraction and planner subgoal updates remain. Context Refinement becomes periodic. Routing-dependent recovery, stall and loop enforcement, and Final Verification are bypassed. \\

Stall and loop handling &
Repeated-action checks and action--result loop checks are disabled. Execution outcomes are not remapped to \texttt{error} or \texttt{stall} from tool failure or repeated no-change evidence. &
Planner subgoal transitions, Step Abstraction, Context Refinement, and Final Verification remain. Runtime recovery feedback for these removed events is not added to planner context. \\

Final Verification &
The verifier call after \texttt{TERMINATE} and its rejection feedback are removed. &
A planner \texttt{TERMINATE} request ends the trajectory directly. All planning, execution, abstraction, routing, refinement, and stall and loop handling before termination remain enabled. \\

\TableBottomRule
\end{tabular}
}
\end{table}

\section{Runtime Control Examples}
\label{app:examples}
\enlargethispage{2\baselineskip}

The runtime-event analysis in Table~\ref{tab:trajectory-control-diagnostics} examines repetition, recovery entry, and rejected termination under the Qwen3.5-9B planner backbone. The six successful OSWorld trajectories below illustrate the model behavior associated with these events and the control mechanisms that guide subsequent execution.

\subsection{Recovery after Repeated Actions}

\begin{taskbox}{01b269ae-2111-4a07-81fd-3fcd711993b0}
\textbf{Instruction.}
Fill all the blank cells in B1:E30 with the value in the cell above it. Finish the work and don't touch irrelevant regions, even if they are blank.

\paragraph{Model Behavior.}
The planner generates three equivalent file-editing commands that produce no visible spreadsheet change. File readback then localizes the unresolved cells to column B. The planner revises the command, reloads Calc, and requests termination before sufficient visual evidence of the target range is available.

\paragraph{Mechanism.}
The persistent subgoal retains the original range and the constraint against modifying irrelevant regions. Step Abstraction incorporates the no-progress and file-readback evidence. L2S revises execution around the unresolved cells instead of restarting the task. Final Verification rejects the premature termination. Subsequent execution refreshes the spreadsheet and obtains sufficient evidence for completion.
\end{taskbox}

\begin{taskbox}{4127319a-8b79-4410-b58a-7a151e15f3d7}
\textbf{Instruction.}
Use terminal command to count all the lines of all php files in current directory recursively, show the result on the terminal

\paragraph{Model Behavior.}
The planner repeatedly uses the Bash channel, which returns the same 54-line result without producing the required visible terminal evidence. It subsequently switches to GUI execution, opens a terminal, and enters the counting command there.

\paragraph{Mechanism.}
The persistent subgoal retains both the recursive counting requirement and the requirement to show the result in the terminal. L2S changes the execution channel while preserving the task stage. Final Verification accepts termination after the visible terminal establishes both the command and its result.
\end{taskbox}

\subsection{Stage-Preserving Recovery across Execution Channels}

\begin{taskbox}{3c8f201a-009d-4bbe-8b65-a6f8b35bb57f}
\textbf{Instruction.}
Download the image from
\url{https://huggingface.co/datasets/xlangai/ubuntu_osworld_file_cache/resolve/main/multi_apps/3c8f201a-009d-4bbe-8b65-a6f8b35bb57f/kingbird.jpeg},
and then use GIMP to compress it to under 600KB as ``compressed.jpeg'' on the Desktop. Resize if needed.

\paragraph{Model Behavior.}
The planner completes the download through Bash and performs the export through the GUI, but repeated GUI attempts produce incorrect filenames. File readback then identifies the candidate files and their sizes, including a valid 115\,KB compressed file.

\paragraph{Mechanism.}
The persistent state preserves the completed download while keeping the compression subgoal active. L2S switches execution back to Bash and normalizes the valid file to the required filename. The accumulated evidence establishes the required location, filename, and size without repeating completed stages.
\end{taskbox}

\begin{taskbox}{c867c42d-a52d-4a24-8ae3-f75d256b5618}
\textbf{Instruction.}
Please assist me in exporting my contacts of Personal Address Book from Thunderbird into contacts.csv file in the desktop and convert it to .xlsx with Libreoffice Calc.

\paragraph{Model Behavior.}
The planner exports the address book through the GUI, producing \texttt{Personal Address Book.csv}. File readback reveals the filename mismatch. The planner then uses Bash to rename the file and completes the conversion in Calc.

\paragraph{Mechanism.}
S2L retains the successful export as completed evidence while keeping the conversion stage active. L2S repairs the filename through Bash without repeating the export and then continues the pending conversion. Bash readback confirms that \texttt{contacts.xlsx} exists on the desktop.
\end{taskbox}

\subsection{Evidence-Grounded Termination}

\begin{taskbox}{1e8df695-bd1b-45b3-b557-e7d599cf7597}
\textbf{Instruction.}
Add a new column named ``Profit'' right next to the 'CGOS' column and calculate the profit for each week by subtracting ``COGS'' from ``Sales'' in that column.

\paragraph{Model Behavior.}
After initial numeric-type and file-state errors, the planner uses a corrected script that completes the edit. Readback confirms the column name and sample calculations. The planner then requests termination before visual confirmation is available.

\paragraph{Mechanism.}
Final Verification rejects the first termination request because the accumulated evidence does not yet establish the visible spreadsheet state. L2S retains the active task stage, resolves the Calc recovery dialog, and verifies the Profit column and its values. Final Verification accepts the subsequent request after the required visual evidence becomes available.
\end{taskbox}

\begin{taskbox}{6ada715d-3aae-4a32-a6a7-429b2e43fb93}
\textbf{Instruction.}
Copy the screenshot 1.png from the desktop to where my cursor is located

\paragraph{Model Behavior.}
The planner requests termination before the available evidence establishes the insertion result. It then inspects the inserted image and its document position. A subsequent termination request occurs while a context menu obscures the relevant content.

\paragraph{Mechanism.}
Final Verification rejects both requests because the available evidence is insufficient to establish the required document state. The persistent image-insertion subgoal keeps execution focused on the unresolved evidence requirement. Subsequent interaction exposes the inserted image and its position, after which Final Verification accepts completion.
\end{taskbox}

\end{document}